\documentclass[10pt,twocolumn,letterpaper]{article}

\usepackage{cvpr}              

\usepackage{lipsum}
\usepackage{graphicx}
\usepackage{amsmath}
\usepackage{amssymb}
\usepackage{multirow}
\usepackage{booktabs}

\usepackage{amsmath,amsfonts,bm,cuted}
\usepackage[ruled,vlined,linesnumbered]{algorithm2e}
\usepackage{graphicx}
\usepackage{tabularx} 
\usepackage{amssymb}
\usepackage{multirow}
\usepackage{subfigure}
\usepackage{xcolor}
\usepackage{dsfont}
\usepackage{upgreek}
\usepackage{url}

\def\eqref#1{equation~\ref{#1}}

\def\1{\bm{1}}

\DeclareMathAlphabet{\mathsfit}{\encodingdefault}{\sfdefault}{m}{sl}
\SetMathAlphabet{\mathsfit}{bold}{\encodingdefault}{\sfdefault}{bx}{n}

\usepackage{times}
 
\normalsize
\usepackage[pagebackref,breaklinks,colorlinks]{hyperref}

\usepackage[capitalize]{cleveref}
\crefname{section}{Sec.}{Secs.}
\Crefname{section}{Section}{Sections}
\Crefname{table}{Table}{Tables}
\crefname{table}{Tab.}{Tabs.}

\hypersetup{pdfstartview={FitH}}

\def\cvprPaperID{4} 
\def\confName{CVPR}
\def\confYear{2022}

\begin{document}

\title{Fine-Grained Visual Classification using Self Assessment Classifier}

\author{Tuong Do$^{1}$, Huy Tran$^{1}$, Erman Tjiputra$^{1}$, Quang D. Tran$^{1}$, Anh Nguyen$^{2}$\\
{$^{1}$AIOZ, Singapore}\\
{$^{2}$University of Liverpool, UK}\\
{\tt\small \{tuong.khanh-long.do, huy.tran, erman.tjiputra,quang.tran\}@aioz.io}\\
{\tt\small anh.nguyen@liverpool.ac.uk}}

\maketitle
\begin{abstract}
Extracting discriminative features plays a crucial role in the fine-grained visual classification task. Most of the existing methods focus on developing attention or augmentation mechanisms to achieve this goal. However, addressing the ambiguity in the top-k prediction classes is not fully investigated. In this paper, we introduce a Self Assessment Classifier, which simultaneously leverages the representation of the image and top-k prediction classes to reassess the classification results. Our method is inspired by continual learning with coarse-grained and fine-grained classifiers to increase the discrimination of features in the backbone and produce attention maps of informative areas on the image. In practice, our method works as an auxiliary branch and can be easily integrated into different architectures. We show that by effectively addressing the ambiguity in the top-k prediction classes, our method achieves new state-of-the-art results on CUB200-2011, Stanford Dog, and FGVC Aircraft datasets. Furthermore, our method also consistently improves the accuracy of different existing fine-grained classifiers with a unified setup. Our source code can be found at: \url{https://github.com/aioz-ai/SAC}.
\end{abstract}

\section{Introduction}
\label{sec:intro}
The fine-grained visual classification task aims to classify images belonging to the same category (e.g., different kinds of birds, aircraft, or flowers). Compared to the ordinary image classification task, classifying fine-grained images is more challenging due to three main reasons: \textit{(i)} large intra-class difference: objects that belong to the same category present significantly different poses and viewpoints; \textit{(ii)} subtle inter-class difference:  objects that belong to different categories might be very similar apart from some minor differences, e.g., the color styles of a bird’s head can usually determine its fine-grained category; \textit{(iii)} limitation of training data:  labeling fine-grained categories usually requires specialized knowledge and a large amount of annotation time. Because of these reasons, it is not a trivial task to obtain accurate classification results by using only the state-of-the-art CNN, such as VGG~\cite{simonyan2014VGG19} and ResNet~\cite{he2016deepResnet152}.

Recent works show that the key solution for fine-grained classification is to find informative regions in multiple object’s parts and extract discriminative features~\cite{Lam_2017_CVPR,wang2018learningDFLCNN,chen2019destructionDCL,zheng2019looking,li2017dynamic,zheng2017learningMultiCNN,huang2021snapmix}. A popular approach to learn object's parts is based on human annotations~\cite{xie2013hierarchical,branson2014bird,zhang2014partRCNN,chai2013symbiotic,gavves2013fine}. However, it is time-consuming to annotate fine-grained regions, hence making this approach impractical.
Some improvements utilize unsupervised or weakly-supervised learning to locate the informative object’s parts~\cite{zheng2017learningMultiCNN,wei2018maskCNNBird} or region of interest bounding boxes~\cite{bilen2016weaklyDec,Fu2017LookCT}. Although this is a promising approach to overcome the problem of manually labeling fine-grained regions, these methods have drawbacks such as low accuracy, costly in training phase/inference phase, or hard to accurately detect separated bounding boxes.

\begin{figure*}
    \centering
    \includegraphics[width=\textwidth, keepaspectratio=true]{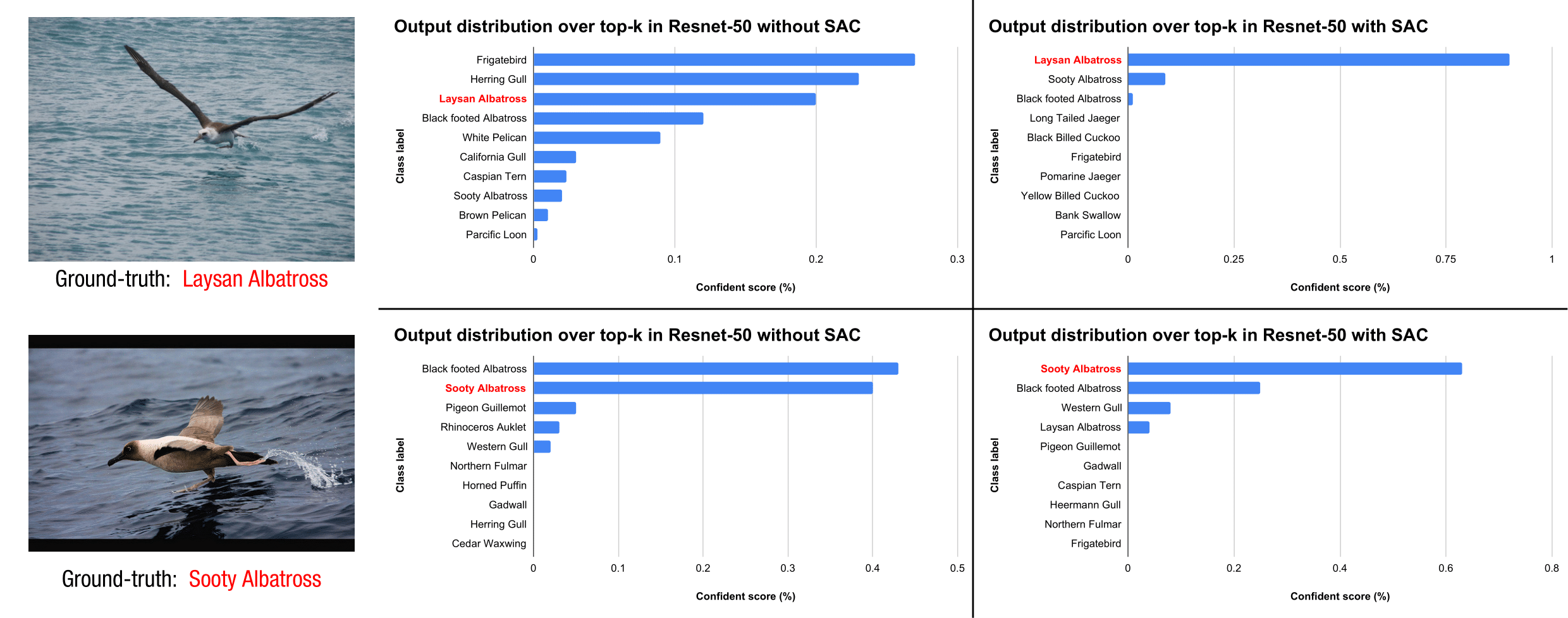}
    \caption{The visualization of the output confident scores over top-10 classes extracted from the softmax layer of the two models: ResNet-50 without SAC and ResNet-50 with SAC.
    The red color class label denotes the ground truth. The visualization indicates that the output distributions extracted from ResNet-50 backbone with our Self Assessment Classifier (SAC) significantly reduce the probability and confident score of wrong classes in the top-10 prediction results. Best viewed in color.
    }
    \label{fig:Uncertainty}
\end{figure*}

In this paper, we introduce a Self Assessment Classifier (SAC) method to address the ambiguity in the fine-grained classification task.
Intuitively, our method is designed to reassess the top-k prediction results and eliminate the uninformative regions in the input image. This helps to decrease the inter-class ambiguity and allows the backbone to learn more discriminative features. During training, our method also produces the attention maps that focus on informative areas of the input image. By integrating into a backbone network, our method can reduce the wrong classification over top-k ambiguity classes. Note that \textit{ambiguity classes} are the results of uncertainty in the prediction that can lead to the wrong classification (See Figure \ref{fig:Uncertainty} for more details). Our contributions can be summarized as follows.
\begin{itemize}
    \item We propose a new self class assessment method that effectively jointly learns the discriminative features and addresses the ambiguity problem in the fine-grained visual classification task. 
    \item  We design a new module that produces attention maps in an unsupervised manner. Based on this attention map, our method can erase the inter-class similar regions that are not useful for classification.
    \item We show that our method can be easily integrated into different fine-grained classifiers to achieve new state-of-the-art results. Our source code and trained models will available for further study.
\end{itemize}

\section{Related Works}
Fine-grained visual classification involves small diversity within the different classes. Typical fine-grained problems, such as differentiating between animal and plant species, drew much attention from researchers. Since background context acted as a distraction in most cases, many pieces of research focus on improving the attentional and localization capabilities of CNN-based algorithms~\cite{wei2018maskCNNBird,zheng2017learningMultiCNN,wang2018learningDFLCNN,chen2019destructionDCL,zhuang2020learningPairwise,wang2020weaklyMixture,gao2020CIN,wang2020graphPro,wang2020progressive,huang2020interpretable,ji2020attention,Zhang2019MGE_CNN,zhang2021MMAL,zhou2020look,zhao2021graph,rao2021counterfactual}. Besides, to focus on the informative regions that could distinguish the species between any two images, many methods relied on annotations of parts location or attributes~\cite{xie2013hierarchical,nguyen2021graph,chai2013symbiotic,le2021global,gavves2013fine}. Specifically, Part R-CNN~\cite{zhang2014partRCNN} and extended R-CNN~\cite{girshick2014richRCNN} detected objects and localized their parts under a geometric prior. Then, these works predicted a fine-grained category from a pose-normalized representation.

In practice, it is expensive to acquire pixel-level annotations of the object’s parts as ground-truth. Thus, methods that require only image-level annotations draw more attention~\cite{hu2019WS_DAN,ge2019weakly3stepmodelling,nguyen2021coarse,yun2019cutmix,huang2021snapmix,xu2021nasoa}. Lin \textit{et al.} proposed the bilinear pooling~\cite{lin2015bilinearpoolingFC} and its improved version~\cite{lin2017improvedPoolingFC}, where two features were combined at each location using the outer product. In~\cite{jaderberg2015spatialSTCNN}, the authors introduce the Spatial Transformer Network to achieve accurate classification performance by learning geometric transformations. In~\cite{hanselmann2020elope}, the authors proposed a lightweight localization module by leveraging global $k$-max pooling. Recently, Joung \textit{et al.}~\cite{joung2021learning} leveraged Canonical 3D Object Representation to cope up with multiple camera viewpoint problem in the fine-grained classification task.

\begin{figure*}[ht]
    \centering
    \includegraphics[width=\textwidth, keepaspectratio=true]{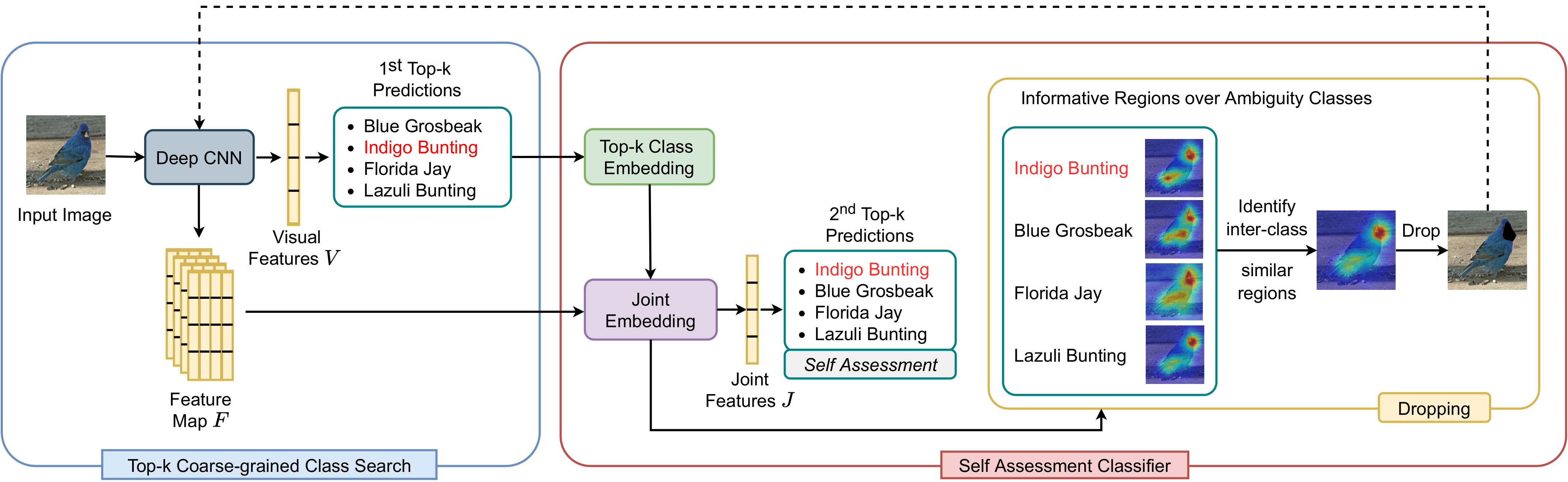}
    \vspace{0.2ex}
    \caption{An overview of our proposed method. The Top-k Coarse-grained Class Search first generates a list of potential top-k prediction candidates using a Deep CNN backbone. The Self Assessment Classifier then reassesses these predictions to improve the fine-grained classification results.
    The red color class label denotes the ground truth. The black color regions denote dropped regions. Dotted line means the image is used for augmentation. Best viewed in color.}
    \vspace{-0.3cm}
    \label{fig:img_class}
\end{figure*}

Many works provided a training routine that maximized the entropy of the output probability distribution for training CNNs~\cite{dubey2018maximumentropy,tran2022light,sun2018MAMC,xie2015hyperregularizer,nguyen2021fl,wang2016miningtripletfinegrained,hu2019WS_DAN,zhang2018mixup,huang2021stochastic}. In~\cite{xie2015hyperregularizer}, the authors exploited a regularization between the fine-grained recognition model and the hyper-class recognition model.
Sun \textit{et al.} proposed  Multiple  Attention  Multiple  Class loss~\cite{sun2018MAMC} that pulled positive features closer to the anchor and pushed negative features away. 
Dubey \textit{et al.} proposed PC~\cite{dubey2018pairwise}, which reduced overfitting by combining the cross-entropy loss with the pairwise confusion loss to learn more discriminative features. 
In~\cite{dubey2018maximumentropy}, by using Maximum-Entropy learning in the context of fine-grained classification, the authors introduced a training routine that maximizes the entropy of the output probability distribution for fine-grained classification task. 
A triplet loss was used in~\cite{wang2016miningtripletfinegrained} to achieve better inter-class separation. 
Hu \textit{et al.}~\cite{hu2019WS_DAN} proposed to use attention regularization loss to focus on attention regions between corresponding local features. More recently, in~\cite{sun2020fine}, diversification block cooperated with gradient-boosting loss had been introduced to maximally separate the highly similar fine-grained classes.

While it is expensive to acquire annotations of object's parts, unsupervised and weakly supervised methods for identifying informative regions are investigated recently. In SCDA~\cite{wei2017selective}, an unsupervised method was introduced to locate the informative regions without using any image label or extra annotation. However, it is less accurate when compared with weakly supervised localization methods, which leveraged image-level super-vision~\cite{hu2019WS_DAN,jaderberg2015spatialSTCNN,ge2019weakly3stepmodelling,lin2015bilinearpoolingFC,jaderberg2015spatialSTCNN,do2021multiple,Kim2017TwoPhaseLF,Zhang2018SelfproducedGF,Choe2019AttentionBasedDL,Bency2016WeaklySL}. To locate the whole object, Zhang \textit{et al.}~\cite{zhang2018adversarial} used Adversarial Complementary Learning which could recognize different object’s parts and discover complementary regions that belong to the same object. Recently, authors in~\cite{wang2020weaklyMixture} used Gaussian  Mixture Model to learn discriminative regions from the image feature maps for fine-grained classification.

All of the above methods do not focus on the ambiguity prediction classes, which is one of the main reasons that causes wrong classifications.
To address this problem, our method is designed to explicitly reduce the effect of the top-k ambiguity prediction classes. Furthermore, our method can effectively learn and produce the attention map in an unsupervised manner. In practice, our method can be easily integrated into different fine-grained classifiers to further improve the classification results.

\section{Methodology}

\subsection{Method Overview}
\label{sec:overview}
We propose two main steps in our method: Top-k Coarse-grained Class Search (TCCS) and Self Assessment Classifier (SAC). TCCS works as a coarse-grained classifier to extract visual features from the backbone. The Self Assessment Classifier works as a fine-grained classifier to to reassess the ambiguity classes and eliminate the non-informative regions. Our SAC has four modules: the Top-k Class Embedding module aims to encode the information of the ambiguity class; the Joint Embedding module aims to jointly learn the coarse-grained features and top-k ambiguity classes; the Self Assessment module is designed to differentiate between ambiguity classes; and finally, the Dropping module is a data augmentation method, designed to erase unnecessary inter-class similar regions out of the input image. Figure \ref{fig:img_class} shows an overview of our approach. 

\subsection{Top-k Coarse-grained Class Search}
\label{subsec:search}

The TCCS takes an image as input. Each input image is passed through a Deep CNN to extract feature map $\textbf{\textit{F}} \in \mathbb{R}^{d_f \times m \times n}$ and the visual feature $\textbf{\textit{V}} \in \mathbb{R}^{d_v}$.
$m, n$, and $d_f$ represent the feature map height, width, and the number of channels, respectively; $d_v$ denotes the dimension of the visual feature $\textbf{\textit{V}}$. In practice, the visual feature $\textbf{\textit{V}}$ is usually obtained by applying some fully connected layers after the convolutional feature map $\textbf{\textit{F}}$.

The visual features $\textbf{\textit{V}}$ is used by the $1^{st}$ classifier, i.e., the original classifier of the backbone, to obtain the top-k prediction results. Assuming that the fine-grained dataset has $N$ classes. The top-k prediction results $\mathds{C}_k = \{C_1,..., C_k\}$ is a subset of all prediction classes $\mathds{C}_N$, with $k$ is the number of candidates that have the $k$-highest confident scores. 

\subsection{Self Assessment Classifier} 
\label{subsec:auxiliary}
Our Self Assessment Classifier takes the image feature $\textbf{\textit{F}}$ and top-k prediction $\mathds{C}_k$ from TCCS as the input to reassess the fine-grained classification results.

\textbf{Top-k Class Embedding.}
\label{subsec:embedding}
The output of TCCS module $\mathds{C}_k$ is passed through top-k class embedding module to output label embedding set $\mathds{E}_k = \{E_1,...E_i,..., E_k\}, i \in \{1,2, ..., k\}, E_i \in \mathbb{R}^{d_{e}}$. This module contains a word embedding layer~\cite{pennington2014glove} for encoding each word in class labels and a GRU~\cite{2014ChoGRU} layer for learning the temporal information in class label names. $d_{e}$ represents the dimension of each class label. It is worth noting that the embedding module is trained end-to-end with the whole model. Hence, the class label representations are learned from scratch without the need of any pre-extracted/pre-trained or transfer learning.

Given an input class label, we trim the input to a maximum of $4$ words. The class label that is shorter than $4$ words is zero-padded. Each word is then represented by a $300$-D word embedding. This step results in a sequence of word embeddings with a size of $4 \times 300$ and denotes as $\hat{E}_i$ of $i$-th class label in $\mathds{C}_k$ class label set. In order to obtain the dependency within the class label name, the $\hat{E}_i$ is passed through a Gated Recurrent Unit (GRU)~\cite{2014ChoGRU}, which results in a $1024$-D vector representation $E_i$ for each input class. Note that, although we use the language modality (i.e., class label name), it is not extra information as the class label name and the class label identity (for calculating the loss) represent the same object category.

\textbf{Joint Embedding.}
\label{subsec:expansion}
This module takes the feature map $\textbf{\textit{F}}$ and the top-k class embedding $\mathds{E}_k$ as the input to produce the joint representation $\textbf{\textit{J}} \in \mathbb{R}^{d_j}$ and the attention map. We first flatten $\textbf{\textit{F}}$ into $(d_f \times f)$, and $\mathds{E}_k$ is into $(d_e \times k)$. The joint representation $\textbf{\textit{J}}$ is calculated using two modalities $\textbf{\textit{F}}$ and $\mathds{E}_k$ as follows
\begin{equation}
\textbf{\textit{J}}^T= \left(\mathcal{T} \times_1 \text{vec}(\textbf{\textit{F}})  \right) \times_2 \text{vec}(\mathds{E}_k)
\label{eq:hypothesis}
\end{equation}
where $\mathcal{T} \in \mathbb{R}^{d_{\textbf{\textit{F}}} \times d_{\mathds{E}_k} \times d_j}$ is a learnable tensor;  $d_{\textbf{\textit{F}}} = (d_f \times f)$; $d_{\mathds{E}_k} = (d_e \times k)$; $\text{vec}()$ is a vectorization operator; $\times_i$ denotes the $i$-mode tensor product.

In practice, the preceding $\mathcal{T}$ is too large and infeasible to learn. Thus, we apply decomposition solutions that reduce the size of $\mathcal{T}$ but still retain the learning effectiveness. 
Inspired by~\cite{Yang2016StackedAN} and~\cite{Kim2018BilinearAN}, we rely on the idea of the unitary attention mechanism. Specifically, let $\textbf{\textit{J}}_p \in \mathbb{R}^{d_j}$ be the joint representation of $p^{th}$ couple of channels where each channel in the couple is from a different input. The joint representation $\textbf{\textit{J}}$ is approximated by using the joint representations of all couples instead of using fully parameterized interaction as in Eq.~\ref{eq:hypothesis}. Hence, we compute $\textbf{\textit{J}}$ as
\begin{equation}
\textbf{\textit{J}} = \sum_p \mathcal{M}_p \textbf{\textit{J}}_p
\label{eq:Unitary}
\end{equation}
Note that in Eq.~\ref{eq:Unitary}, we compute a weighted sum over all possible couples. The $p^{th}$ couple is associated with a scalar weight $\mathcal{M}_p$. The set of $\mathcal{M}_p$ is called as the attention map $\mathcal{M}$, where $\mathcal{M} \in \mathbb{R}^{f \times k}$.

There are $f \times k $ possible couples over the two modalities. The representation of each channel in a couple is $\textbf{\textit{F}}_{i}, \left(\mathds{E}_k\right)_{j}$, where $i \in  [1,f], j \in  [1,k]$, respectively. The joint representation $\textbf{\textit{J}}_p$ is then computed as follow
\begin{equation}
\textbf{\textit{J}}_p^T= \left(\mathcal{T}_{u} \times_1 \textbf{\textit{F}}_{i} \right)\times_2 \left(\mathds{E}_k\right)_{j}
\label{eq:couple_compute}
\end{equation}
where $\mathcal{T}_{u} \in \mathbb{R}^{d_f \times d_e \times d_j}$ is the learning tensor between channels in the couple.

From Eq.~\ref{eq:Unitary}, we can compute the attention map $\mathcal{M}$ using the reduced parameterized bilinear interaction over the inputs $\textbf{\textit{F}}$ and $\mathds{E}_k$. The attention map is computed as
\begin{equation}
\mathcal{M} = \text{softmax}\left(\left(\mathcal{T}_\mathcal{M} \times_1  \textbf{\textit{F}} \right) \times_2 \mathds{E}_k \right)
\label{eq:bi_attmap}
\end{equation}
where $\mathcal{T}_\mathcal{M} \in \mathbb{R}^{d_f \times d_e}$ is the learnable tensor. 

By integrating Eq.~\ref{eq:couple_compute} and Eq.~\ref{eq:bi_attmap} into  Eq.~\ref{eq:Unitary}, the joint representation $\textbf{\textit{J}}$ can be rewritten as
\begin{equation}
\textbf{\textit{J}}^T= \sum_{i=1}^{f}\sum_{j=1}^{k} \mathcal{M}_{ij} \left( \left( \mathcal{T}_{u} \times_1 \textbf{\textit{F}}_{i}\right) \times_2 \left(\mathds{E}_k\right)_{j} \right)
\label{eq:transformed_hypo}
\end{equation}

It is also worth noting from Eq.~\ref{eq:transformed_hypo} that to compute $\textbf{\textit{J}}$, instead of learning the  large tensor $\mathcal{T} \in \mathbb{R}^{d_{F} \times d_{\mathds{E}_k} \times d_j}$ in Eq.~\ref{eq:hypothesis}, we now only need to learn two smaller tensors  $\mathcal{T}_{u} \in \mathbb{R}^{d_{f} \times d_{e} \times d_j}$  in Eq.~\ref{eq:couple_compute} and $\mathcal{T}_\mathcal{M} \in \mathbb{R}^{d_f \times d_e}$ in Eq.~\ref{eq:bi_attmap}.

\textbf{Self Assessment.}
\label{subsec:assessment}
The joint representation $\textbf{\textit{J}}$ from the Joint Embedding module is used as the input in the Self Assessment step to obtain the $2^{nd}$ top-k predictions $\mathds{C}'_k$.
Note that $\mathds{C}'_k = \{C'_1,..., C'_k\}$. Intuitively, $\mathds{C}'_k$ is the top-k classification results after self-assessment. This module is a fine-grained classifier that produces the $2^{nd}$ predictions to reassess the ambiguity classification results.

In practice, we train a classifier using a Cross-Entropy loss function to obtain the top-k prediction results.
Through back-propagation, the feature map $\textbf{\textit{F}}$ from the Deep CNN backbone and the joint features $\textbf{\textit{J}}$ are expected to be more useful for fine-grained classification. By jointly train both the coarse-grained and fine-grained classifiers, our method can increase the discriminative of feature maps. Therefore, we can reduce the the ambiguity classes and improve the overall fine-grained classification accuracy. 

Inspired by~\cite{hu2018weaklyBAN,gebru2017fine}, the contribution of the coarse-grained and fine-grained classifier is calculated by
\begin{equation}
\text{Pr}(\small{\rho} = \small{\rho}_i) = \alpha \text{Pr}_1(\small{\rho} = \small{\rho}_i) + (1- \alpha) \text{Pr}_2(\small{\rho} = \small{\rho}_i)
\label{eq:weighted_sum}
\end{equation}
where $\alpha$ is the trade-off hyper-parameter $\left(0 \leq \alpha \leq 1\right)$. $\text{Pr}_1(\small{\rho} = \small{\rho}_i), \text{Pr}_2(\small{\rho} = \small{\rho}_i)$ denote the prediction probabilities for class $\small{\rho}_i$, from the coarse-grained and fine-grained classifiers, respectively. 

It should be emphasized that in our design, three modules Top-k Class Embedding, Joint Embedding, and Self Assessment have to be utilized \textit{simultaneously} to produce the final classification predictions.

\textbf{Dropping.}
\label{subsec:dropping}
Although the attention map $\mathcal{M}$ can identify the correlation between the image regions and each ambiguity class, unnecessary regions may still exist and affect the fine-grained classification results. To encourage the attention maps to pay more attention to the informative regions which are discriminate between ambiguity classes, we introduce a Dropping mechanism. Our dropping method leverages the current attention map $\mathcal{M}$ to eliminate regions which have high correlation scores, \textit{i.e.}, inter-class similar regions. It then recomputes the attention distribution between the remaining regions and the top-k classes. 
It should be emphasized that the eliminated regions are not backgrounds. They are foregrounds and may also be essential for classification. However, these regions are not discriminative enough for the fine-grained classification task. 

\SetKwInput{KwInput}{Input}               
\SetKwInput{KwOutput}{Output}             
\begin{algorithm}
\DontPrintSemicolon
  \KwInput{Network model, Input image $\textbf{\textit{I}}$}
  \KwOutput{Cropped area $\textbf{\textit{I}}_c$}
  Pass $\textbf{\textit{I}}$ through Self Assessment Classifier to extract attention map $\mathcal{M}$.
  
  Sum up the correlation scores over all top-k class labels in $\mathcal{M}$ and then apply bilinear interpolation to get the correlation scores of all regions in the input image.
  
  Extract maximum correlation score $\gamma_{\text{max}}$ from $\mathcal{M}$, which denotes the highest correlation score in the attention map $\mathcal{M}$. 
  
  Compute minimum correlation score $\gamma_{\text{min}}=0.1 * \gamma_{\text{max}}$. Only regions which have higher correlation scores than $\gamma_{\text{min}}$ are kept.
  
  Update attention map $\mathcal{M}$ by using Eq.~\ref{eq:updated_atm}. Correlation scores, which are smaller than $\gamma_{\text{min}}$, are set to zero. 
  \begin{equation}
   \mathcal{M} = \left( \mathcal{M} \geq \gamma_{\text{min}}\right) * \mathcal{M}
   \label{eq:updated_atm}
  \end{equation}
  
  To identify the cropped area, we compute top left point $\textbf{\textit{P}}_1(x_1,y_1)$ and bottom right point $\textbf{\textit{P}}_2(x_2,y_2)$ by using Eq.~\ref{eq:crop}.
  \begin{equation}
    \begin{aligned}
    \label{eq:crop}
     &x_1 = \text{min}\left(\text{IdxRowSelected}\left(\mathcal{M}\right)\right)\\
     &y_1 = \text{min}\left(\text{IdxColSelected}\left(\mathcal{M}\right)\right)\\
     &x_2 = \text{max}\left(\text{IdxRowSelected}\left(\mathcal{M}\right)\right)\\
     &y_2 = \text{max}\left(\text{IdxColSelected}\left(\mathcal{M}\right)\right)\\
    \end{aligned}
  \end{equation}
  where IdxRowSelected$\left(\mathcal{M}\right)$ and IdxColSelected$\left(\mathcal{M}\right)$ extract the row index list and the column index list of correlation values which are higher than zero in $\mathcal{M}$, respectively.
  
  \textbf{return} $\textbf{\textit{I}}_c$ identified by $\textbf{\textit{P}}_1$ and $\textbf{\textit{P}}_2$
\caption{Region Localization}
\label{alg:localization}
\end{algorithm}

Specifically, for each $c$-th ambiguity class, we first obtain the dropping mask $\mathcal{M}_{D_c} \in \mathbb{R}^{f}$ from the attention map $\mathcal{M}_c$ by setting $i$-th element $\mathcal{M}_{c_i}$ which is larger than threshold $(d_{\phi} * \text{max}(\mathcal{M}))$  to 0, and others to 1. Eq.~\ref{eq:masking} shows this condition. Note that $\mathcal{M}_c \in \mathbb{R}^{f}$ is the $c$-th attention map extracted from the attention map $\mathcal{M}$; $c \in \left[1,k \right]$ where $k$ is the number of ambiguity classes.
\begin{equation}
\mathcal{M}_{D_{c_i}} = \begin{cases} 
   0, & \text{if } \mathcal{M}_{c_i} > d_{\phi}* \text{max}(\mathcal{M}) \\
   1,       & \text{otherwise}
  \end{cases}
\label{eq:masking}
\end{equation}

The dropping masks of ambiguity classes are then integrated into the feature map to remove unnecessary regions. The final feature map $\textbf{\textit{F}}' \in \mathbb{R}^f$ is calculated as follow
\begin{equation}
\textbf{\textit{F}}' = \left(\mathcal{M}_{D_1} \vee \mathcal{M}_{D_2} \vee ... \vee \mathcal{M}_{D_k}\right) \odot \textbf{\textit{F}}
\label{eq:recompted_feature_map}
\end{equation}
where $\odot$ denotes Hadamard product.

In practice, the Self Assessment Classifier plays an important role during training to learn discriminate features and resolve the ambiguity in the top-k prediction classes. During testing, inspired by~\cite{hu2019WS_DAN}, we can also reuse the attention map $\mathcal{M}$ of our Self Assessment Classifier to remove the unnecessary background from the image. We describe this cropping localization step in Algorithm \ref{alg:localization}. 

\section{Experiment}
\subsection{Experimental Setup}
\label{subset:dataset}
\textbf{Dataset.} We evaluate our method on three popular fine-grained datasets: CUB-200-2011~\cite{wah2011caltechCUB}, Stanford Dogs~\cite{khosla2011novelDogs} and FGVC Aircraft~\cite{maji2013fineAirCraft}. Table \ref{tab:dataset} shows the statistic of these datasets. Note that we do not use any extra bounding box/part annotations in all of our experiments. 

\begin{table}[h]
\centering
\small
\setlength{\tabcolsep}{0.45em} 
{\renewcommand{\arraystretch}{1.2}
\begin{tabular}{c|c|c|c|c}
\hline
\textbf{Dataset}       & {\textbf{\begin{tabular}[c]{@{}c@{}} Target\end{tabular}}}          & \textbf{\begin{tabular}[c]{@{}c@{}} \# Category\end{tabular}}  & \textbf{\begin{tabular}[c]{@{}c@{}} \# Train\end{tabular}} & \textbf{\begin{tabular}[c]{@{}c@{}} \# Test\end{tabular}}\\ \hline
CUB-200-2011~\cite{wah2011caltechCUB} & Bird &$200$ &$5,994$ &$5,794$   \\ \hline
Stanford Dogs~\cite{khosla2011novelDogs}  & Dog &$120$ &$12,000$ &$8,580$   \\ \hline
FGVC-Aircraft~\cite{maji2013fineAirCraft}  & Aircraft &$100$ &$6,667$ &$3,333$  \\ \hline
\end{tabular}
}
\vspace{0.25cm}
\caption{Three fine-grained visual classification datasets that are used in our experiments.
}
\label{tab:dataset}
\end{table}

\textbf{Implementation.}
All experiments are conducted on an NVIDIA Titan V GPU with 12GB RAM. The model is trained using Stochastic Gradient Descent with a momentum of 0.9. The maximum number of epochs is set at 80; the weight decay equals 0.00001, and the mini-batch size is 12. Besides, the initial learning rate is set to 0.001, with exponential decay of 0.9 after every two epochs. Based on validation results, the number of top-k ambiguity classes is set to 10, while the parameters $d_{\phi}$, $\alpha$ are set to $0.1$ and $0.5$, respectively. The dimension of the embedding of top-k classes and the joint representation between top-k class embedding and feature map is set to $1024$. The training and testing time of our method depends on the backbone it is being integrated to. However, it is not $1.1$ times slower compared to the original baseline. 

\textbf{Baseline.} To validate the effectiveness and generalization of our method, we integrate it into $7$ different deep networks, including two popular Deep CNN backbones, Inception-V3~\cite{szegedy2016inceptionv3} and ResNet-50~\cite{he2016deepResnet152}; and five fine-grained classification methods: WS~\cite{hu2018weaklyBAN}, DT~\cite{Cui2018LargeSFG}, WS\_DAN~\cite{hu2019WS_DAN}, MMAL~\cite{zhang2021MMAL}, and the recent transformer work ViT~\cite{conde2021exploring}. It is worth noting that we only add our Self Assessment Classifier into these works, other setups and hyper-parameters for training are kept unchanged when we compare with original codes. Although our proposed method is integrated into different approaches and tested on different datasets, we use the same parameter setup as described above in all experiments.

\subsection{Fine-grained Classification Results}
\label{subset:sota}

Table \ref{tab:abl_generalization} summarises the contribution of our Self Assessment Classifier (SAC) to the fine-grained classification results of different methods on three datasets CUB-200-2011, Stanford Dogs, and FGVC Aircraft. This table clearly shows that by integrating SAC into different classifiers, the fine-grained classification results are consistently improved. In particular, we observe an average improvement of  $+1.3$, $+1.2$, and $+1.2$ in the CUB-200-2011, Stanford Dogs, and FGVC Aircraft datasets, respectively. Our SAC shows a clear improvement when being integrated into Inception-V3 and ResNet-50 on three datasets. It is more challenging to improve the results of existing fine-grained classification methods, however, we still achieve consistent improvement when integrating SAC into WS~\cite{hu2018weaklyBAN}, DT~\cite{Cui2018LargeSFG}, WS\_DAN~\cite{hu2019WS_DAN}, MMAL~\cite{zhang2021MMAL}, and ViT~\cite{conde2021exploring} fine-grained classifiers. Table \ref{tab:abl_generalization} illustrates that by integrating with recent fine-grained classifiers, our proposed method achieves new state-of-the-art results in both CUB-200-2011, Stanford Dogs, and FGVC Aircraft datasets.

To conclude, our proposed method shows a clear and consistent improvement on different classifiers and different fine-grained datasets. In practice, our method can be easily integrated into new state-of-the-art classifiers to further improve the results. The parameter setting for our method is also simple as we obtain good results with one setup across different classifiers on different datasets. 

\begin{table}[!t]
\begin{center}
\footnotesize
\setlength{\tabcolsep}{0.4em} 
{\renewcommand{\arraystretch}{1.2}
\begin{tabular}{l|c|c|c} 
\hline
\multirow{3}{*}{ \textbf{Methods}}                              & \multicolumn{3}{c}{\textbf{Acc (\%)}}                         \\ 
\cline{2-4}
                                                                & \textbf{\begin{tabular}[c]{@{}c@{}}CUB-200-\\2011 \end{tabular}}         & \textbf{\begin{tabular}[c]{@{}c@{}}Stanford\\ Dogs \end{tabular}}  & \textbf{\begin{tabular}[c]{@{}c@{}}FGVC\\ Aircraft \end{tabular}}      \\ 
\hline
RA\_CNN~\cite{Fu2017LookCT}                                                            &$85.4$                      &$87.3$                      & $88.4$                      \\ 

MAMC~\cite{sun2018MAMC}                                                            &$86.5$                      &$85.2$                      & \_                      \\ 

PC~\cite{dubey2018pairwise}                                                            &$86.9$                      &$83.8$                      & $89.2$                      \\ 
MC~\cite{chang2020mutualchannel}                                 &$87.3$                      &\_                      &$92.9$                       \\ 
DCL~\cite{chen2019destructionDCL}                                                          &$87.8$                      &\_                      & $93.0$                      \\
ACNet~\cite{ji2020attention}                                                          &$88.1$                      &\_                      & $92.4$                      \\
DF-GMM~\cite{wang2020weaklyMixture}                                                          &$88.8$                      &\_                      & $93.8$                      \\ 
API-Net~\cite{zhuang2020learningPairwise}                                                          &$90.0$                      &$90.3$                      & $93.9$                      \\ 
GHORD~\cite{zhao2021graph}                                 &$89.6$                      &\_                      &$94.3$                       \\
CAL~\cite{rao2021counterfactual}                                 &$90.6$                      &\_                      &$94.2$                       \\
Parts Models~\cite{ge2019weakly3stepmodelling}                                 &$90.4$                      &$93.9$                      &\_                       \\ 
\hline
Inception-V3~\cite{szegedy2016inceptionv3}                                                    &$83.7$                      & $85.1$                     &$87.4$                      \\
Inception-V3-SAC                                                   & $85.3 (+1.6)$                     &  $86.8 (+1.7)$                      & $89.2(+1.8)$                   \\
\hline
ResNet-50~\cite{CGNLNetwork2018}                                                       &$86.4$                      & $86.1$                   & $90.3$                     \\ 
ResNet-50-SAC                                                      &$88.3 (+1.9)$                     & $87.4 (+1.3)$                     & $92.1 (+1.8)$                      \\
\hline
WS~\cite{hu2018weaklyBAN}                             &$88.8$                      &$91.4$                      &$92.3$                       \\
WS-SAC                              & $89.9 (+1.1)$                      & $92.5 (+1.1)$                    & $93.2(+0.9)$                     \\ 
\hline
DT~\cite{Cui2018LargeSFG}                              &$89.2$                      &$88.0$                      &$90.7$                       \\ 

DT-SAC                             &$90.1 (+0.9)$                      & $88.8 (+0.8)$                    & $91.9 (+1.2)$                     \\ 

\hline
MMAL~\cite{zhang2021MMAL}                                                         &$89.6$                      &$90.6$                      &$94.7$                       \\ 
MMAL-SAC                              &$90.8 (+1.2)$                       &$91.6 (+1.0)$                      &$\textbf{95.5} (+0.8)$                      \\ 
\hline
WS\_DAN~\cite{hu2019WS_DAN}                                                         &$89.4$                      &$92.2$                      &$93.0$                       \\ 
WS\_DAN-SAC                              &${91.1} (+1.7)$                       &$93.1 (+0.9)$                      &$93.9 (+0.9)$                      \\ 
\hline
ViT~\cite{conde2021exploring}                                                         &$91.0$                      &$93.2$                      &$92.1$                       \\ 
ViT-SAC                              &$\textbf{91.8} (+0.8)$                       &$\textbf{94.5} (+1.3)$                      &$93.1 (+1.0)$                      \\ 
\hline
\hline
\textbf{Avg. Improvement}  &$+1.3$                      &$+1.2$                      &$+1.2$                       \\ 
\hline
\multicolumn{1}{l}{}                                            & \multicolumn{1}{l}{} & \multicolumn{1}{l}{} & \multicolumn{1}{l}{} 
\end{tabular}
}
\end{center}
\vspace{-0.35cm}
\caption{Contribution of our Self Assessment Classifier (SAC) on fine-grained classification results when we integrate it into different methods.
}
\label{tab:abl_generalization}
\vspace{-0.3cm}
\end{table}

\subsection{Module Study Analysis}
\label{subset:abl}

\textbf{Module Contribution}. Table \ref{tab:abl_module_evaluate} shows the contribution of each module when we integrate our method into ResNet-50 and WS\_DAN on the CUB-200-2011 dataset. We note that, based on the design of our SAC, three modules Top-k Class Embedding, Joint Embbeding, and Self Assessment must be utilized simultaneously. We denote them as the Auxiliary Classifier in Table \ref{tab:abl_module_evaluate}. By adopting the Auxiliary Classifier, both accuracies of ResNet-50 and WS\_DAN are improved by $+0.7\%$ and $+0.6\%$, respectively. This result indicates that the backbone under SAC has learned informative representations for dealing with fine-grained ambiguity classes. We also observe that applying the Dropping module further improves the performance on both ResNet-50 and WS-DAN. This confirms that by removing inter-class similar regions, the backbone can learn more discriminative information, hence improving the classification results. Table \ref{tab:abl_module_evaluate} also shows that all components in our Self Assessment Classifier contribute significantly to the final results. 
 
\begin{table}
\begin{center}
\small
\setlength{\tabcolsep}{0.16em} 
{\renewcommand{\arraystretch}{1.2}
\begin{tabular}{c|c|c|c|c|c|c}
\hline
\multicolumn{2}{c|}{\textbf{Backbone}}                                                                     & \checkmark & \checkmark                                              & \checkmark                                              & \checkmark                                            & \checkmark                                              \\ \hline
\multicolumn{2}{c|}{\textbf{Auxiliary Classifier}}                                                     &      & \checkmark                                              &                                                & \checkmark                                                                & \checkmark                                              \\ \hline
\multicolumn{2}{c|}{\textbf{Dropping}}                                                                     &      &                                                   & \checkmark                                              & \checkmark                                                               & \checkmark                                              \\ \hline
\multicolumn{2}{c|}{\textbf{Localization}} &      &                                                   &                                                   &                                                                     & \checkmark                                              \\ \hline
\multicolumn{2}{c|}{\textbf{ResNet-50 Backbone}}                                                                   & \begin{tabular}[c]{@{}c@{}}86.4\\ (+0.0)\end{tabular} & \begin{tabular}[c]{@{}c@{}}87.1\\ (+0.7)\end{tabular} & \begin{tabular}[c]{@{}c@{}}87.6\\ (+1.2)\end{tabular} & \begin{tabular}[c]{@{}c@{}}88.0\\ (+1.6)\end{tabular} &  \begin{tabular}[c]{@{}c@{}}88.3\\ (+1.9)\end{tabular} \\ \hline
\multicolumn{2}{c|}{\textbf{WS\_DAN Backbone}}                                                                 & \begin{tabular}[c]{@{}c@{}}89.4\\ (+0.0)\end{tabular} & \begin{tabular}[c]{@{}c@{}}90.0\\ (+0.6)\end{tabular} & \begin{tabular}[c]{@{}c@{}}90.3\\ (+0.9)\end{tabular} & \begin{tabular}[c]{@{}c@{}}90.7\\ (+1.3)\end{tabular} &  \begin{tabular}[c]{@{}c@{}}91.1\\ (+1.7)\end{tabular} \\ \hline
\end{tabular}
}
\end{center}
\vspace{-0.1cm}
\caption{The performance (\% Acc) of each module in our method the on CUB-200-2011 test set. ResNet-50 and WS\_DAN are used as the backbones in this experiment. 
}
\label{tab:abl_module_evaluate}

\end{table}

\begin{table}[!t]
\centering
\small
\setlength{\tabcolsep}{0.3em} 
{\renewcommand{\arraystretch}{1.2}
\begin{tabular}{c|c|c|c}
\hline
{\textbf{\begin{tabular}[c]{@{}c@{}}$\alpha$\\(coarse-grained)\end{tabular}}} & {\textbf{\begin{tabular}[c]{@{}c@{}}($1-\alpha$)\\ (fine-grained)\end{tabular}}} & {\textbf{\begin{tabular}[c]{@{}c@{}}ResNet-50 \\ +SAC\end{tabular}}} & {\textbf{\begin{tabular}[c]{@{}c@{}}WS\_DAN \\ +SAC\end{tabular}}} \\ \hline
0.5                                                                                           & 0.5                                                                                     & 88.3 & 91.1                                                               \\ \hline
0.7                                                                                           & 0.3                                                                                     & 88.1 & 91.0                                                               \\ \hline
0.3                                                                                           & 0.7                                                                                     & 88.3 & 91.2                                                               \\ \hline
0.9                                                                                           & 0.1                                                                                     & 88.0 & 91.0                                                               \\ \hline
0.1                                                                                           & 0.9                                                                                     & 87.8 & 90.9                                                               \\ \hline
\end{tabular}
}
\caption{The performance of ResNet-50 + SAC and WS\_DAN + SAC under different values of parameter $\alpha$, which controls the contribution of coarse-grained classifier and fine-grained classifier.}
\label{tab:weightsum_analysis}
\vspace{-0.3 cm}
\end{table}

\textbf{Coarse vs. Fine-grained Classifier Analysis.}
In this work, we consider that both the coarse-grained classifier and the fine-grained classifier are equally important. In practice, we can control the contribution of each classifier by changing the parameter $\alpha$ in Eq.~\ref{eq:weighted_sum}. In all our experiments, the $\alpha$ parameter is set to $0.5$, which means both coarse and fine-grained classifier contributes equally to the classification results. For completeness, Table \ref{tab:weightsum_analysis} is provided to validate the effect of this parameter using ResNet-50 and WS\_DAN on the CUB-200-2011 dataset. This table demonstrates that by fine-tuning the $\alpha$ parameter, the fine-grained classification results can be slightly improved. However, we can see that the final classification results do not depend too much on this $\alpha$ parameter.

\textbf{Complexity Analysis.} Table \ref{tab:complexity} shows the efficiency of each module of SAC and its complexity indicated by the GPU speed and the number of parameters during inference process, when we integrate SAC into ResNet-50 \cite{he2016deepResnet152} and WS\_DAN \cite{hu2019WS_DAN} on CUB-200-2011 dataset. By leveraging Auxiliary Classifier and Dropping module, both inference time and the number of parameters of ResNet-50 and WS\_DAN are unchanged. These results imply that SAC increases the performance without affecting the computational cost of backbones. 

\begin{table}[]
\centering
\small
\setlength{\tabcolsep}{0.8em} 
{\renewcommand{\arraystretch}{1.2}
\begin{tabular}{c|c|c}
\hline
\textbf{\#Top-k classes} & \textbf{\begin{tabular}[c]{@{}c@{}} ResNet-50 + SAC\end{tabular}}          & \textbf{\begin{tabular}[c]{@{}c@{}} WS\_DAN + SAC\end{tabular}} 
\\\hline
$2$   & $87.2$ & $89.4$    \\
\hline
$5$   & $88.4$ & $89.6$    \\ \hline
$10$  & $88.3$ & $91.1$    \\ \hline
$20$  & $87.7$ & $90.7$    \\ \hline
$50$  & $85.9$ & $87.2$    \\ \hline
\end{tabular}
}
\caption{The effect of different numbers of top-k classes.
}
\label{tab:topk_analysis}
\vspace{0.1cm}
\end{table}

From Table \ref{tab:complexity}, we can see that both the network parameters and inference time are increased when we use Localization. However, the increased complexity is an acceptable trade-off to achieve better results. 

\begin{table*}[!ht]
\begin{center}
\setlength{\tabcolsep}{0.24em} 
{\renewcommand{\arraystretch}{1.3}
\begin{tabular}{c|c|c|c|c|c|c}
\hline
\multicolumn{2}{c|}{\textbf{Backbone}}                                                                     & \checkmark & \checkmark                                              & \checkmark                                                   & \checkmark                                              & \checkmark                                              \\ \hline
\multicolumn{2}{c|}{\textbf{\begin{tabular}[c]{@{}c@{}}Auxiliary\\Classifier\end{tabular}}}                                                     &      & \checkmark                                              &                                              & \checkmark                                                             & \checkmark                                              \\ \hline
\multicolumn{2}{c|}{\textbf{Dropping}}                                                                     &      &                                                   & \checkmark                                                              &   \checkmark                                            & \checkmark                                              \\ \hline
\multicolumn{2}{c|}{\textbf{Localization}}    &      &                                                   &                                                   &                                                                                & \checkmark                                              \\ \hline
\multirow{3}{*}{\textbf{\begin{tabular}[c]{@{}c@{}}ResNet-50\\ Backbone\end{tabular}}}                                                                                     & \textit{\#Parameters (M)} & 25.6     & 25.6                                                  &25.6                                                   & 25.6                                                                                              & 38.4   \\ \cline{2-7} 
                                                                                    & \textit{\begin{tabular}[c]{@{}c@{}}GPU Speed\\ (s/sample)\end{tabular}} 
     &\begin{tabular}[c]{@{}c@{}}$0.009$\\ $\pm 0.0013$ \end{tabular}
     &\begin{tabular}[c]{@{}c@{}}$0.009$\\ $\pm 0.0013$\end{tabular}         &\begin{tabular}[c]{@{}c@{}}$0.009$ \\ $\pm 0.0013$\end{tabular}          &\begin{tabular}[c]{@{}c@{}}$0.009$ \\ $\pm 0.0013$\end{tabular}                   &\begin{tabular}[c]{@{}c@{}}$0.017$ \\ $\pm 0.0022$\end{tabular}\\ \hline
\multirow{3}{*}{\textbf{\begin{tabular}[c]{@{}c@{}}WS\_DAN\\ Backbone\end{tabular}}} 
                                                                                    & \textit{\#Parameters (M)} &29.8      & 29.8                                                  & 29.8                                                   & 29.8                                                                                            &49.0  \\ \cline{2-7} 
                                                                                    & \textit{\begin{tabular}[c]{@{}c@{}}GPU Speed\\ (s/sample)\end{tabular}} 
     &\begin{tabular}[c]{@{}c@{}}$0.121$\\ $\pm 0.0110$ \end{tabular}
     &\begin{tabular}[c]{@{}c@{}}$0.121$\\ $\pm 0.0110$\end{tabular}         &\begin{tabular}[c]{@{}c@{}}$0.121$ \\ $\pm 0.0110$\end{tabular}          &\begin{tabular}[c]{@{}c@{}}$0.121$ \\ $\pm 0.0110$\end{tabular}                  &\begin{tabular}[c]{@{}c@{}}$0.201$ \\ $\pm 0.0120$\end{tabular}\\ \hline
\end{tabular}
}
\end{center}
\caption{Performance and complexity of each module of SAC evaluated on CUB-200-2011 test set. ResNet-50 and WS\_DAN is used for baselines as well as the backbones.
}
\label{tab:complexity}
\end{table*}

\subsection{Number of Top-k Classes Analysis.}
The accuracy of our proposed method depends on the top-k prediction classes extracted dynamically by the coarse-grained classifier. If the coarse-grained classifier has poor performance and the top-k classes value is set at a small number, there may be no ground truth class in any in top-k predictions. In this case, our fine-grained classifier only penalizes the wrong cases. Therefore, the fine-grained classifier can not improve the accuracy of the network. Table~\ref{tab:topk_analysis} shows the effect of the number of top-k ambiguity classes on the classification results in our method. From this table, we can see that if the number of top-k classes is set to a small number, our improvement is minimal. In practice, we recommend setting this parameter to a relatively big number to avoid this problem. We choose $k=10$ in all of our experiments with different methods and datasets.

\subsection{Dropping Analysis} 
\textbf{Dropping Comparison.} 
Table \ref{tab:augmentation_comparison} illustrates the performance of our ambiguity class based dropping, in comparison with other augmentation methods. Weakly Supervised learning model (WS~\cite{hu2018weaklyBAN}) and Weakly Supervised Data Augmentation Network (WS\_DAN~\cite{hu2019WS_DAN}) are used as the baselines and also the backbones in our experiment. It is worth noting that we follow the hyper-parameters and evaluation metrics mentioned in~\cite{hu2019WS_DAN} for a fair comparison. Specifically, our proposed Dropping module improves the accuracy of both WS and WS\_DAN by $2.1$ and $0.9$, respectively.
These values indicate that the dropping module in our method successfully removes ambiguity regions between top-k classes and achieves competitive results with other augmentation methods.

\begin{table}
\begin{center}
\small
\setlength{\tabcolsep}{0.3em}
{\renewcommand{\arraystretch}{1.2}
\begin{tabular}{l|c}
\hline
\textbf{Method}                                                                                   & \begin{tabular}[c]{@{}l@{}}Acc (\%)\end{tabular} \\ \hline
WS~\cite{hu2018weaklyBAN} (Backbone)                                                                                     & 86.4                           \\ \hline
WS + Random Cropping   ~\cite{devries2017improvedcutout}                                                                              & 86.8                         \\
WS + Random Dropping~\cite{devries2017improvedcutout}                                                                    & 86.7                      \\
\begin{tabular}[c]{@{}l@{}}WS + Random Cropping \& Dropping~\cite{devries2017improvedcutout}    \end{tabular}                 & 87.2                             \\ \hline
WS + Attention Cropping~\cite{hu2019WS_DAN}                                                               & 87.8                             \\ 
WS + Attention Dropping~\cite{hu2019WS_DAN}                                                              & 87.4                             \\ 
\begin{tabular}[c]{@{}l@{}}WS + Attention Cropping \& Dropping~\cite{hu2019WS_DAN}\end{tabular} & 88.4                     \\ \hline
WS + Dropping (ours)                                                                                & 88.5                     \\
\hline
\hline
WS\_DAN~\cite{hu2019WS_DAN} (Backbone)                                                                                & 89.4                           \\
WS\_DAN + Dropping (ours)                                                                           & \textbf{90.3}                     \\
\hline
\end{tabular}
}
\end{center}
\caption{Comparison between our Dropping method and different augmentation techniques such as Random Dropping~\cite{devries2017improvedcutout} and Attention Cropping~\cite{hu2019WS_DAN}.
}
\label{tab:augmentation_comparison}
\end{table}

\textbf{Dropping Threshold.}
Dropping is an essential step in our method as well as in other fine-grained works. In practice, dropped regions can enhance the discriminative between the top-k classes, but they also may hurt the original discrimination between the ground truth and the remaining classes. We can control this problem in our method by using the hyper-parameter $d_{\phi}$. Table \ref{tab:dphi_Analysis} shows the result of our method under different values of $d_{\phi}$ on ResNet-50 and WS\_DAN baselines. If $d_{\phi}$ is set too small, the effectiveness of the dropping module is not enough for improving the overall performance. If $d_{\phi}$ is set too large, the dropping module may mistakenly drop the discriminative features for different images that belong to the same class, hence reducing the accuracy. Through empirical results, we observe that dropping threshold $d_{\phi} = 0.1$ shows the consistent performance across CNN backbones and fine-grained datasets.

\begin{table}[]
\centering
\small
\setlength{\tabcolsep}{0.2em}
{\renewcommand{\arraystretch}{1.2}
\begin{tabular}{c|c|c}
\hline
\textbf{$d_{\phi}$ (dropping threshold)}       & {\textbf{\begin{tabular}[c]{@{}c@{}} ResNet-50+SAC\end{tabular}}}          & \textbf{\begin{tabular}[c]{@{}c@{}} WS\_DAN+SAC\end{tabular}} \\ \hline
$0.05$ & 87.6 & 90.5   \\ \hline
$0.1$  & 88.3 & 91.1   \\ \hline
$0.2$  & 88.4 & 90.9   \\ \hline
$0.5$  & 83.9 & 87.0   \\ \hline
\end{tabular}
}
\vspace{0.25cm}
\caption{The performance (\% Acc) of ResNet-50 + SAC and WS\_DAN + SAC with different values of dropping threshold $d_{\phi}$. 
}
\label{tab:dphi_Analysis}
\end{table}

\subsection{Visualization}
\begin{figure}[]
\centering
\includegraphics[scale=0.129]{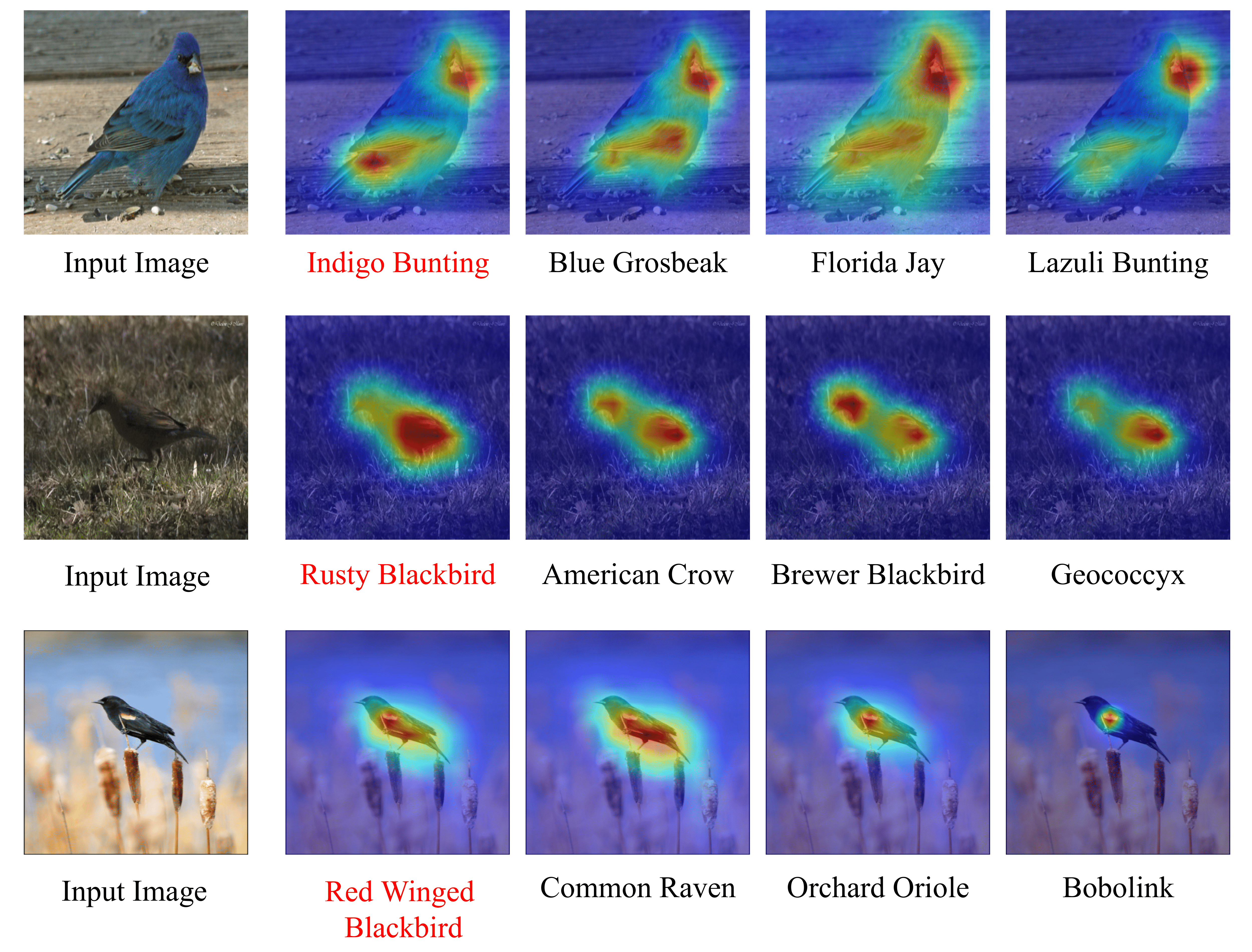}
\caption{The visualization of the attention map between image feature maps and different ambiguity classes from our method. The red-colored class label denotes that the prediction is matched with the ground-truth. Best viewed in color.
}
\label{fig:bird_att_vis}
\end{figure}

\textbf{Qualitative results for Attention Maps.}
Figure \ref{fig:bird_att_vis} illustrates the visualization of attention maps between image feature maps and each ambiguity class. The visualization indicates that by employing our Self Assessment Classifier, each fine-grained class focuses on different informative regions. 

\begin{figure}[]
  \centering
    \subfigure[]{\includegraphics[width=0.185\linewidth, height=0.6\linewidth]{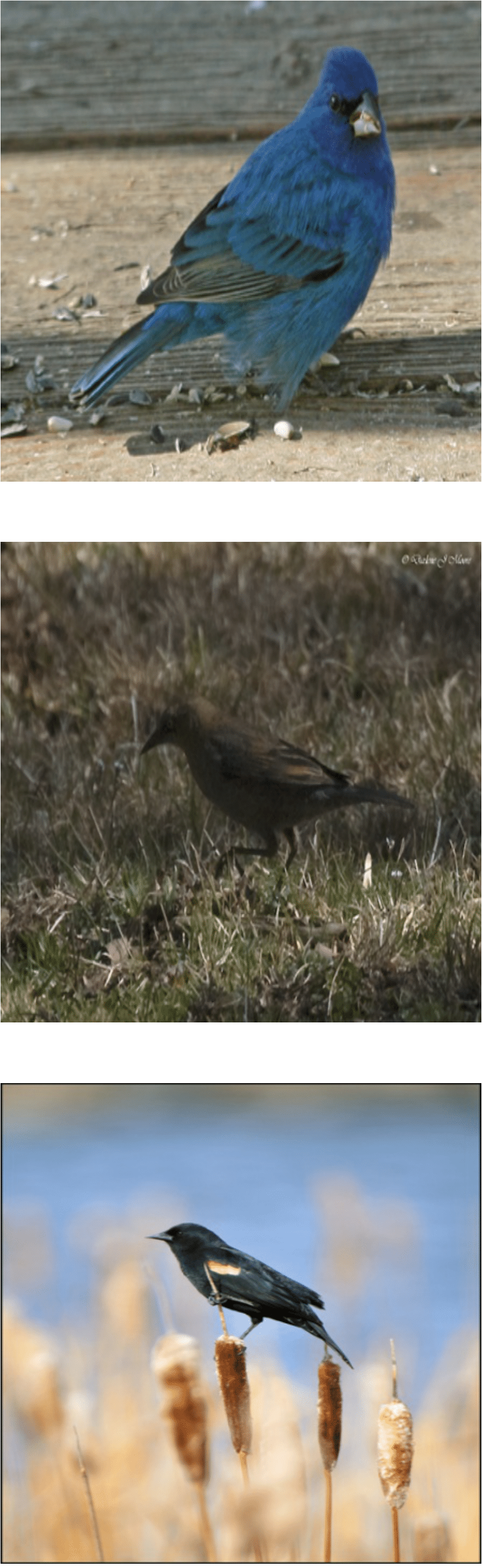}}
    \subfigure[]{\includegraphics[width=0.185\linewidth, height=0.6\linewidth]{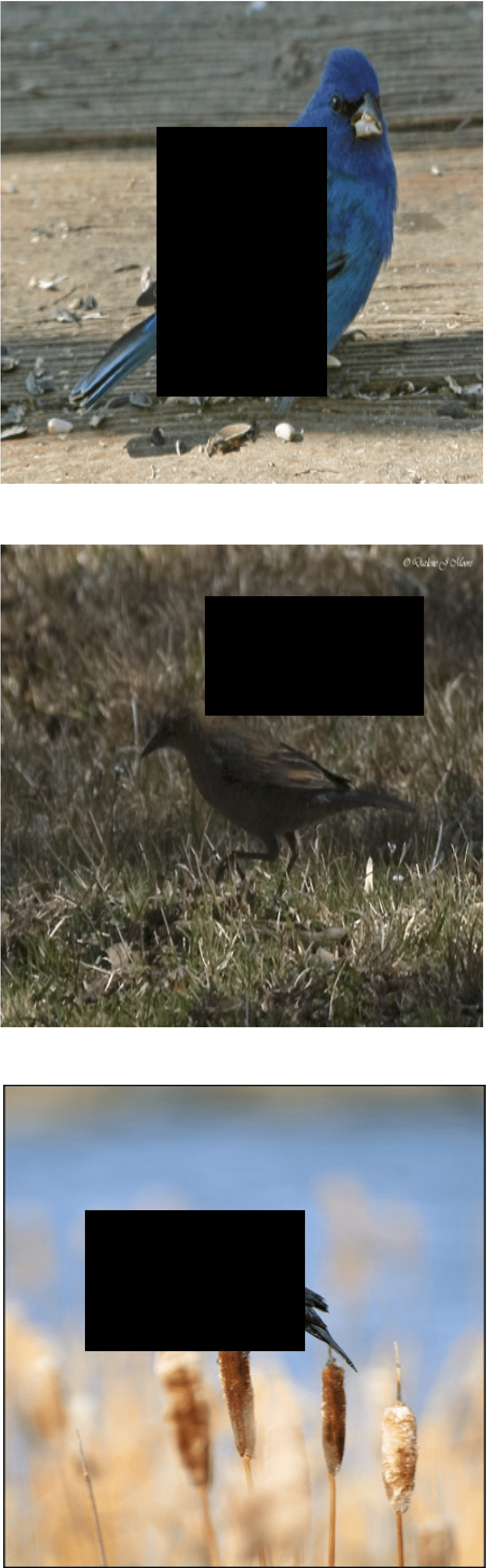}}
    \subfigure[]{\includegraphics[width=0.185\linewidth, height=0.6\linewidth]{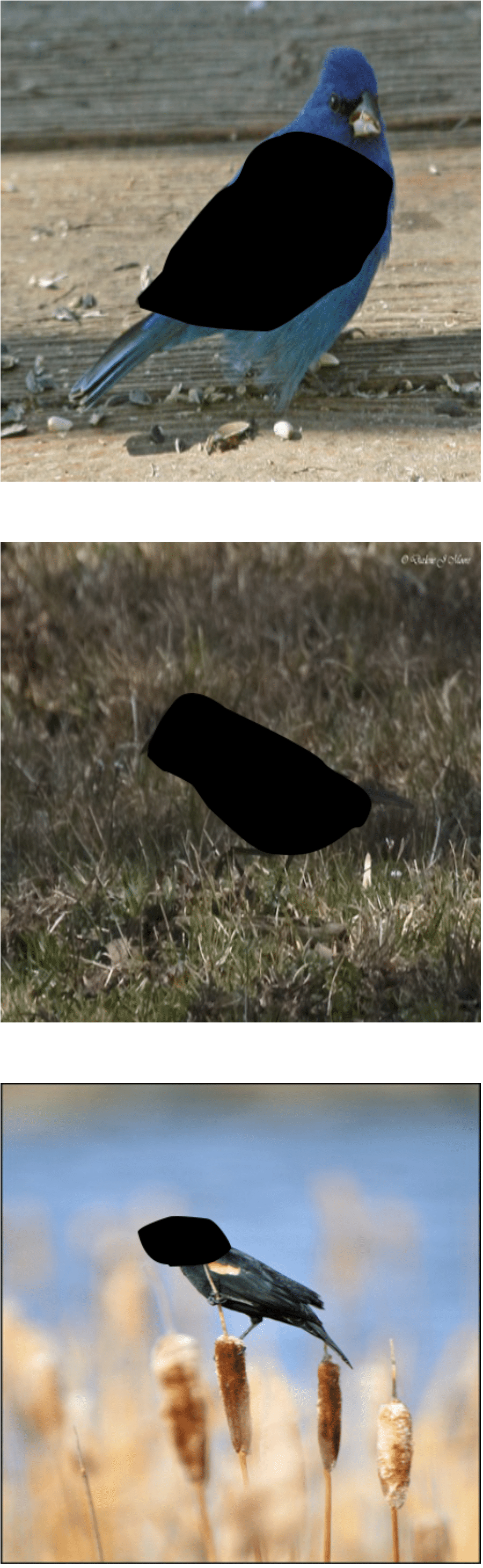}}
    \subfigure[]{\includegraphics[width=0.185\linewidth, height=0.6\linewidth]{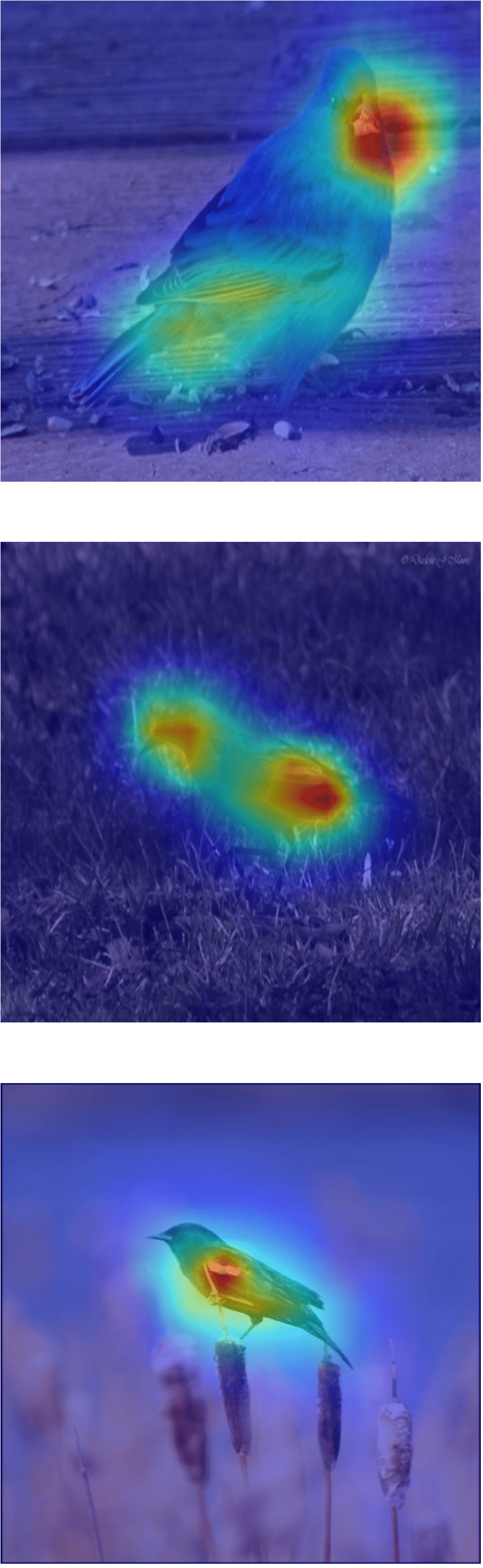}}
    \subfigure[]{\includegraphics[width=0.185\linewidth, height=0.6\linewidth]{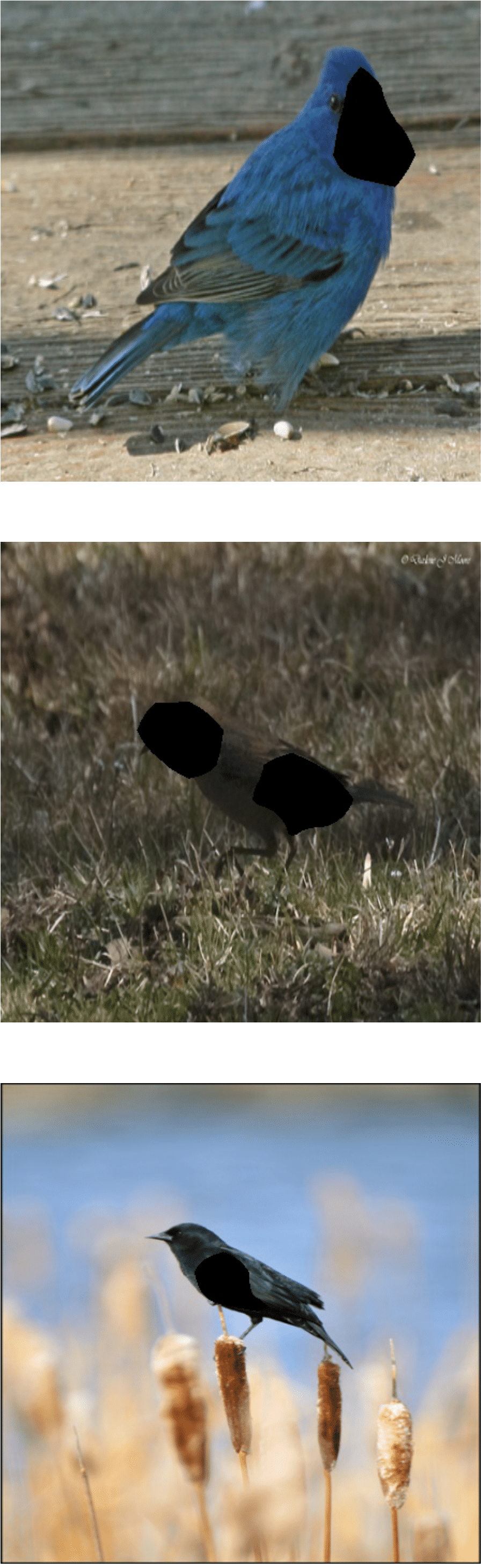}}
    \vspace{0.3ex}
 \caption{The visualization of different dropping techniques. (a) Input image, (b) Random Dropping~\cite{devries2017improvedcutout}, (c) Attention Dropping~\cite{hu2019WS_DAN}, (d) The distribution maps from our Dropping module,  (e) Our Dropping. Best viewed in color.}
 \label{fig:bird_dropping_vis}
\end{figure}

\textbf{Qualitative results between Dropping methods.}
In Figure \ref{fig:bird_dropping_vis}, we illustrate the visual comparison between Attention Dropping~\cite{hu2019WS_DAN}, Random Dropping~\cite{devries2017improvedcutout}, and our SAC dropping methods. Unlike Random Dropping, which can erase the entire object or just the background, and Attention Dropping, which can erase discriminative informative regions to increase generalization, our proposed dropping method can efficiently identify and erase the informative regions that lead to incorrect classifications.

\begin{figure}[!ht]
   \centering
\resizebox{\linewidth}{!}{
\setlength{\tabcolsep}{2pt}
\begin{tabular}{cccc}
\shortstack{\includegraphics[width=0.33\linewidth]{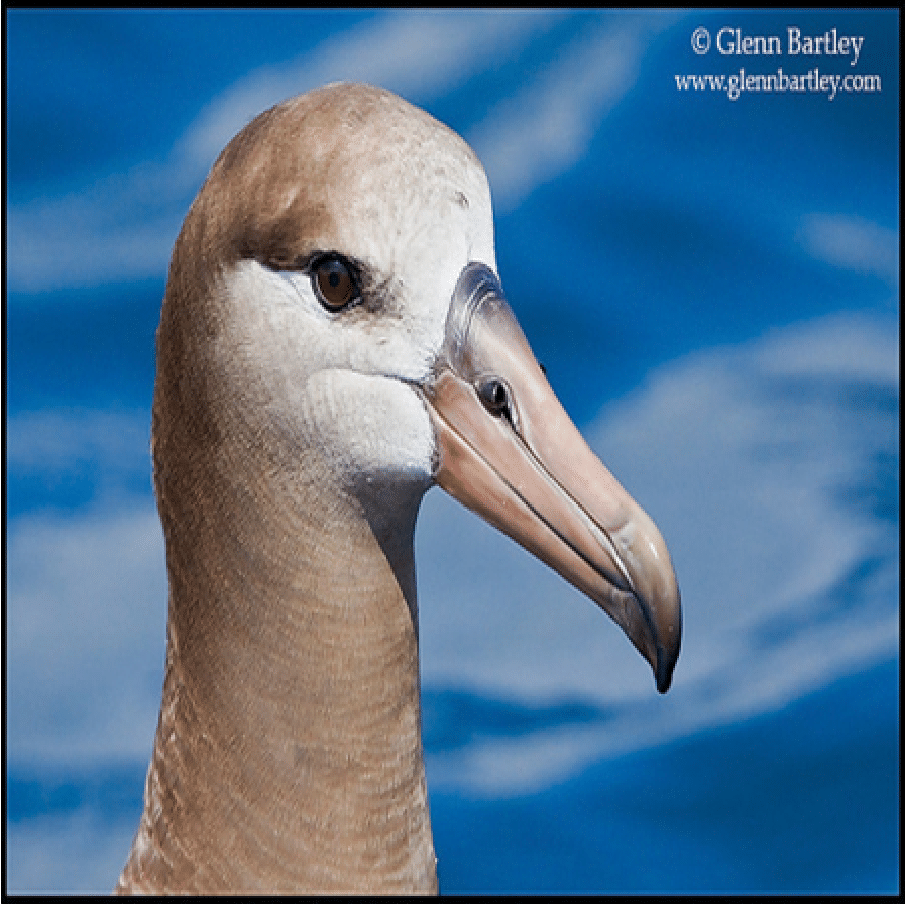}}&
\shortstack{\includegraphics[width=0.33\linewidth]{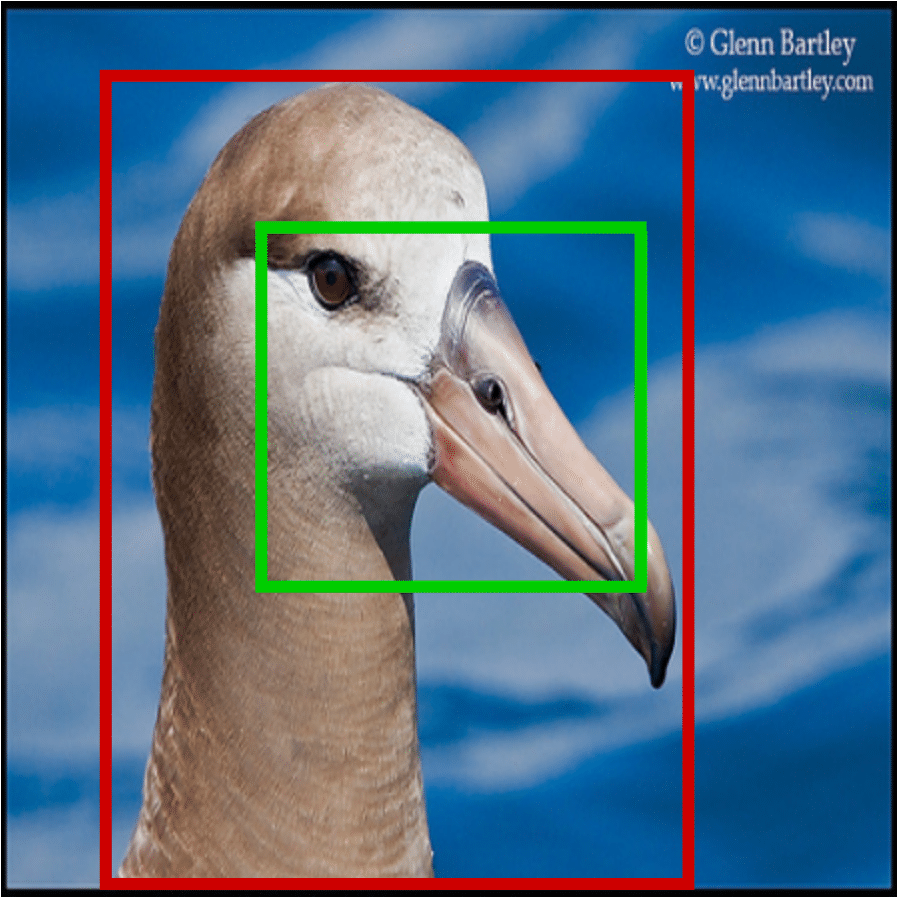}}&
\shortstack{\includegraphics[width=0.33\linewidth]{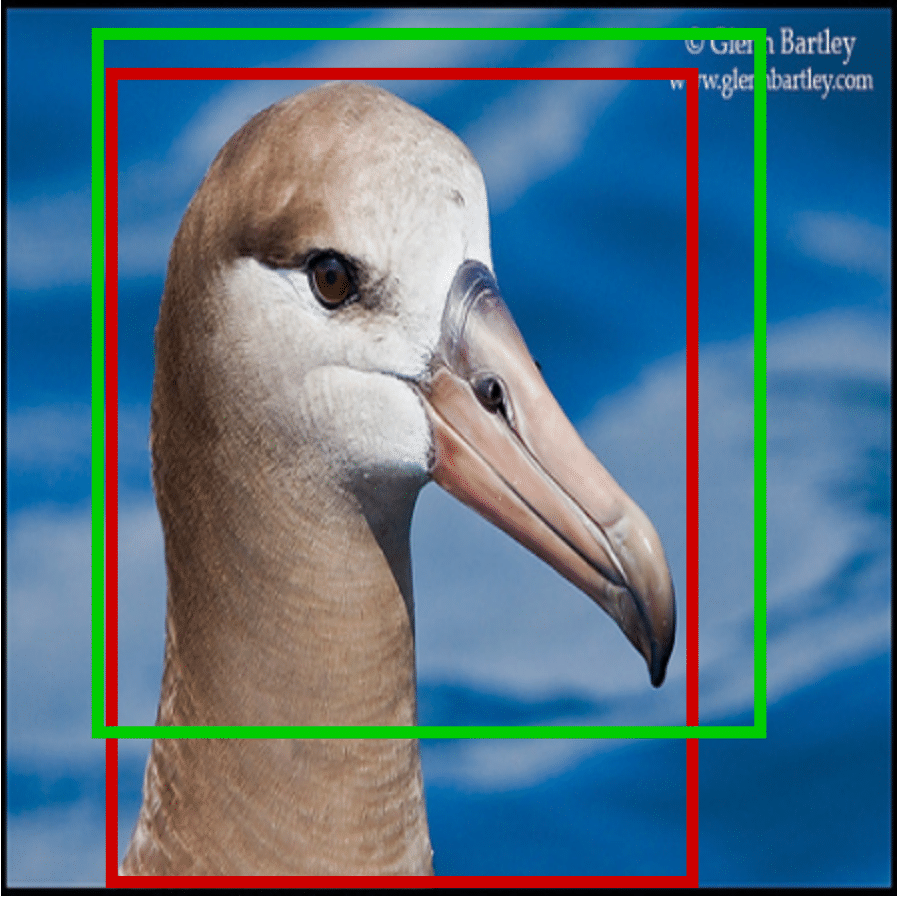}}&
\shortstack{\includegraphics[width=0.33\linewidth]{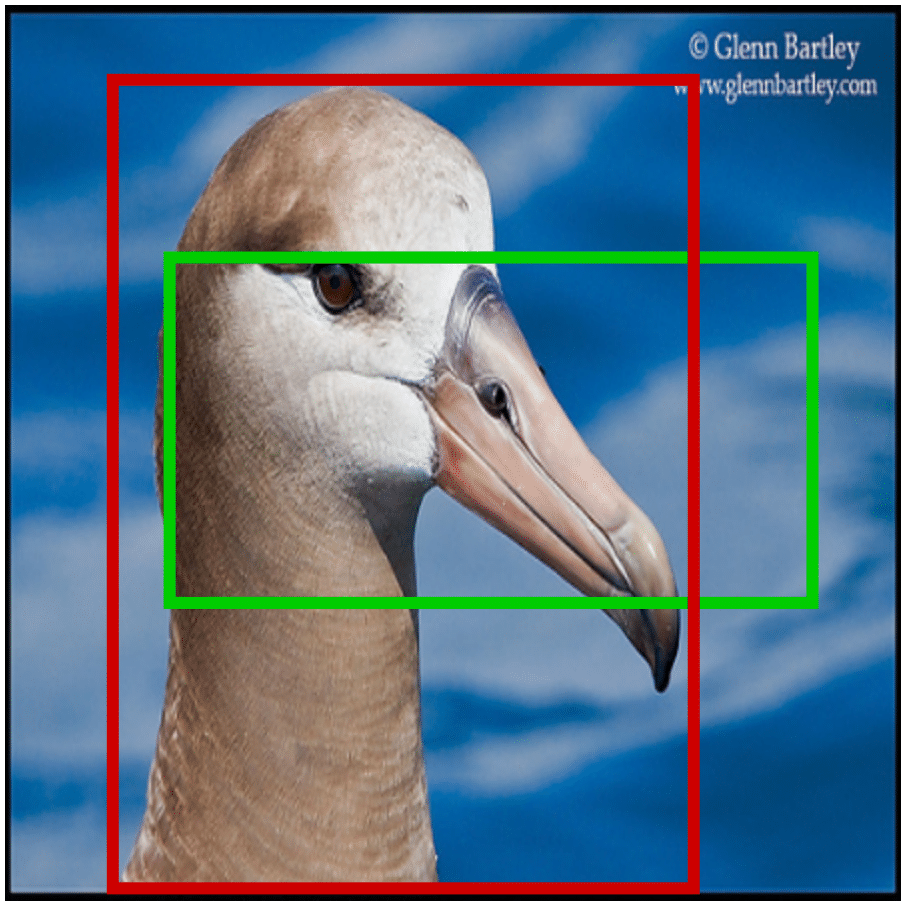}}\\[1pt]
\shortstack{\includegraphics[width=0.33\linewidth]{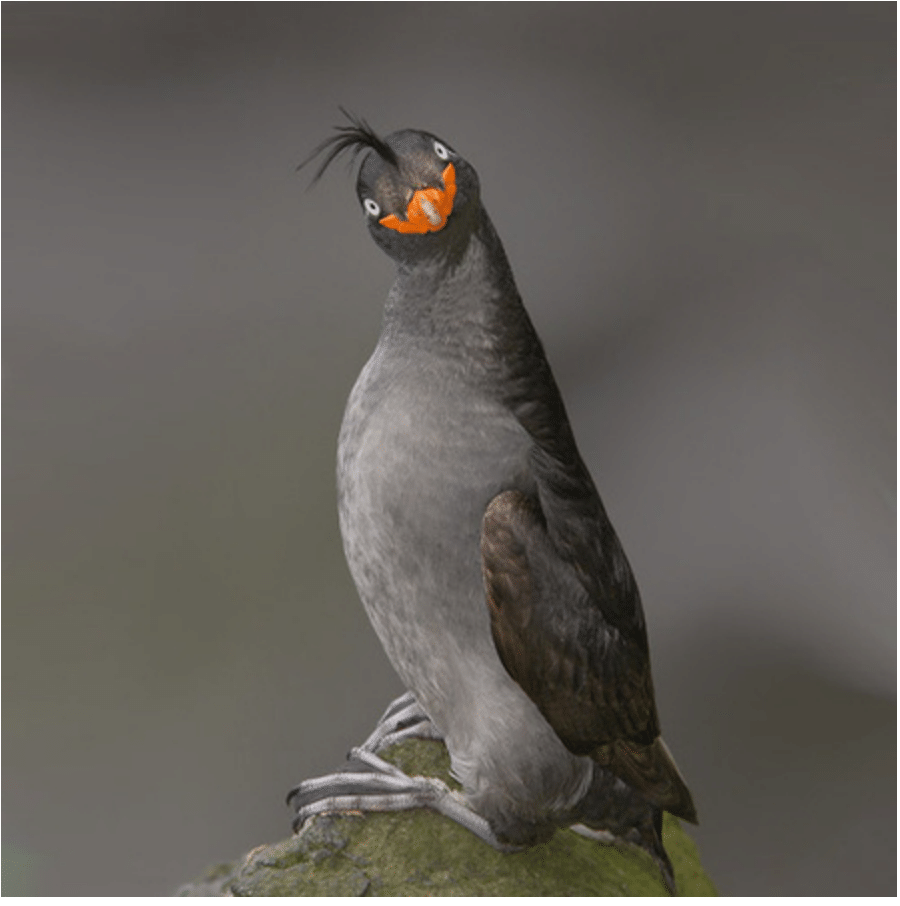}}&
\shortstack{\includegraphics[width=0.33\linewidth]{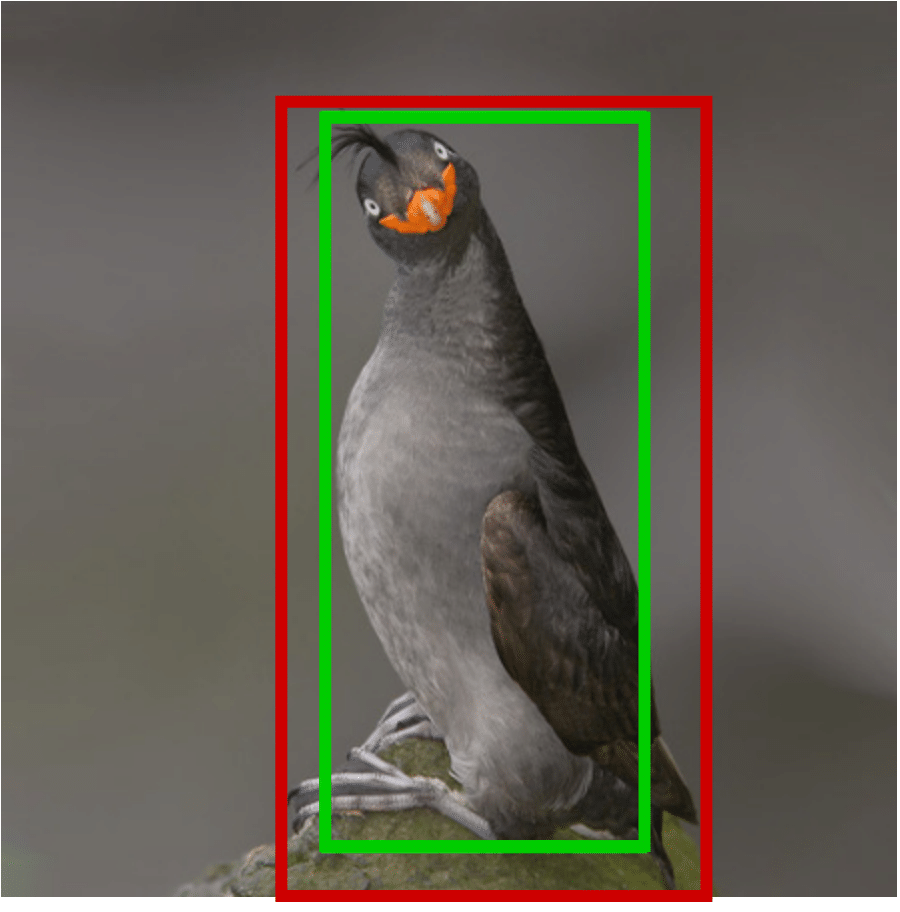}}&
\shortstack{\includegraphics[width=0.33\linewidth]{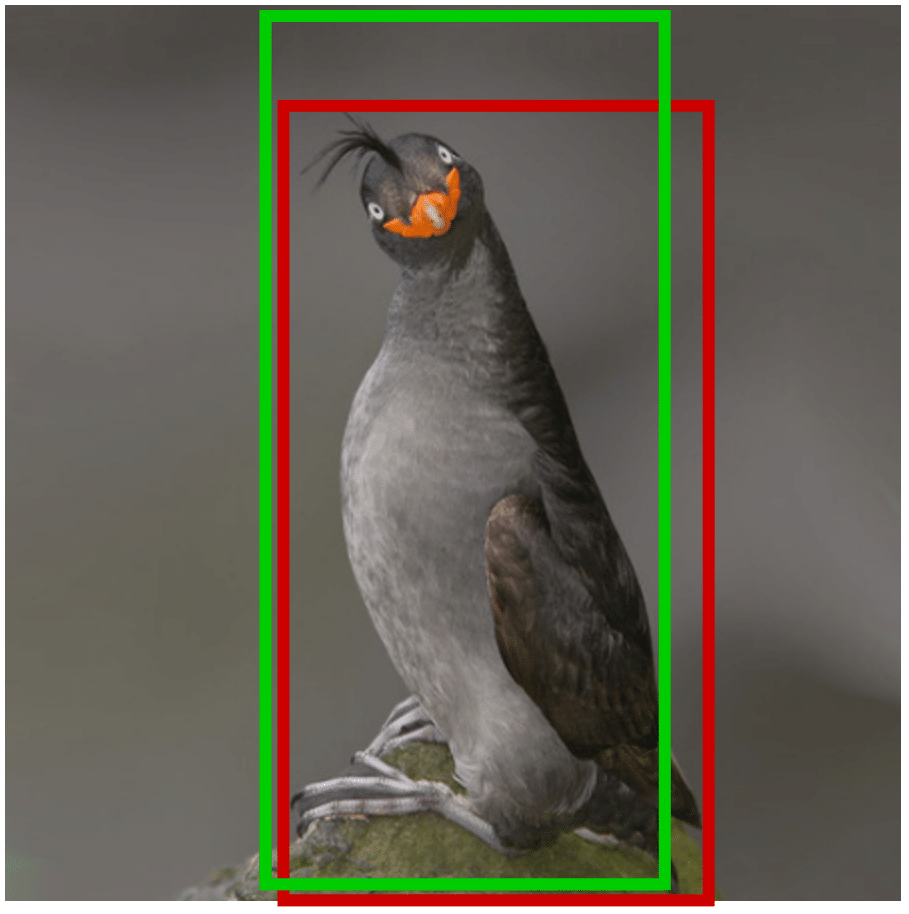}}&
\shortstack{\includegraphics[width=0.33\linewidth]{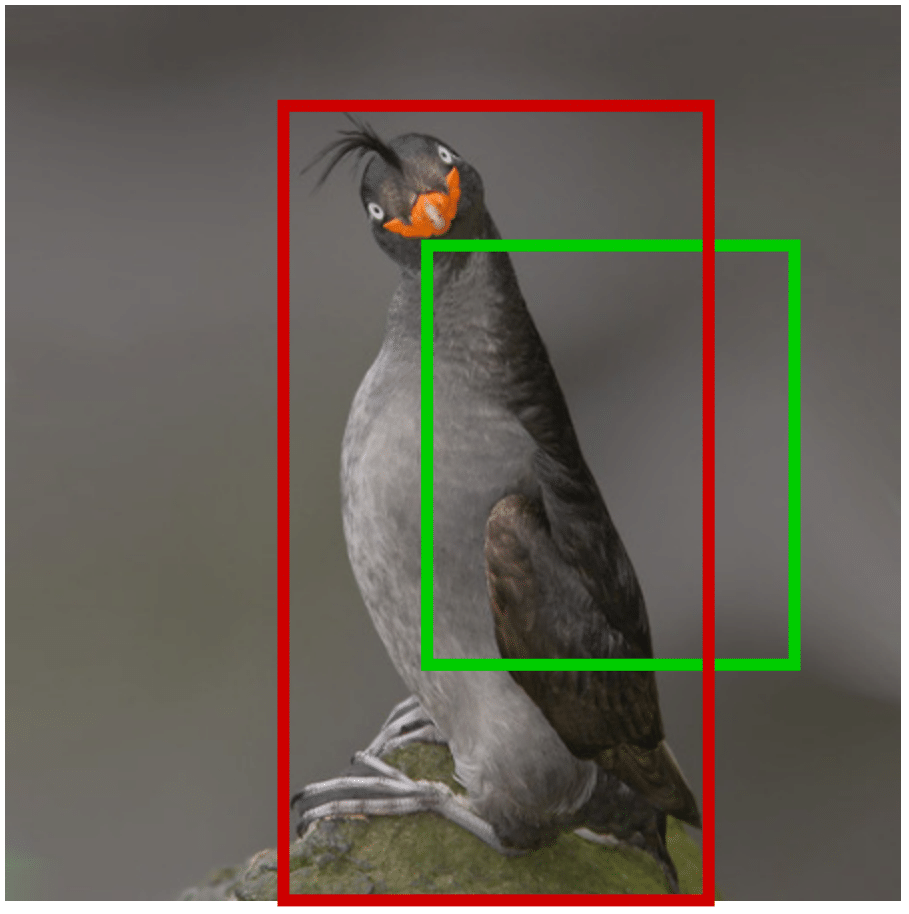}}\\[1pt]
\shortstack{\includegraphics[width=0.33\linewidth]{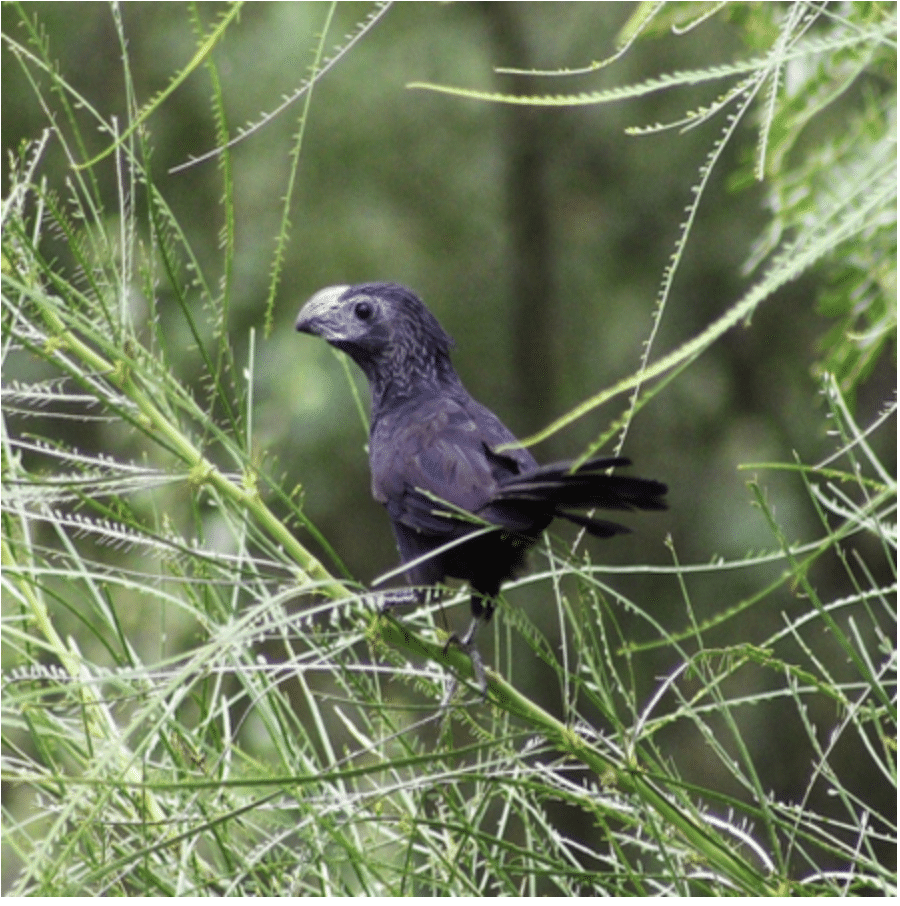}\\  (a)}&
\shortstack{\includegraphics[width=0.33\linewidth]{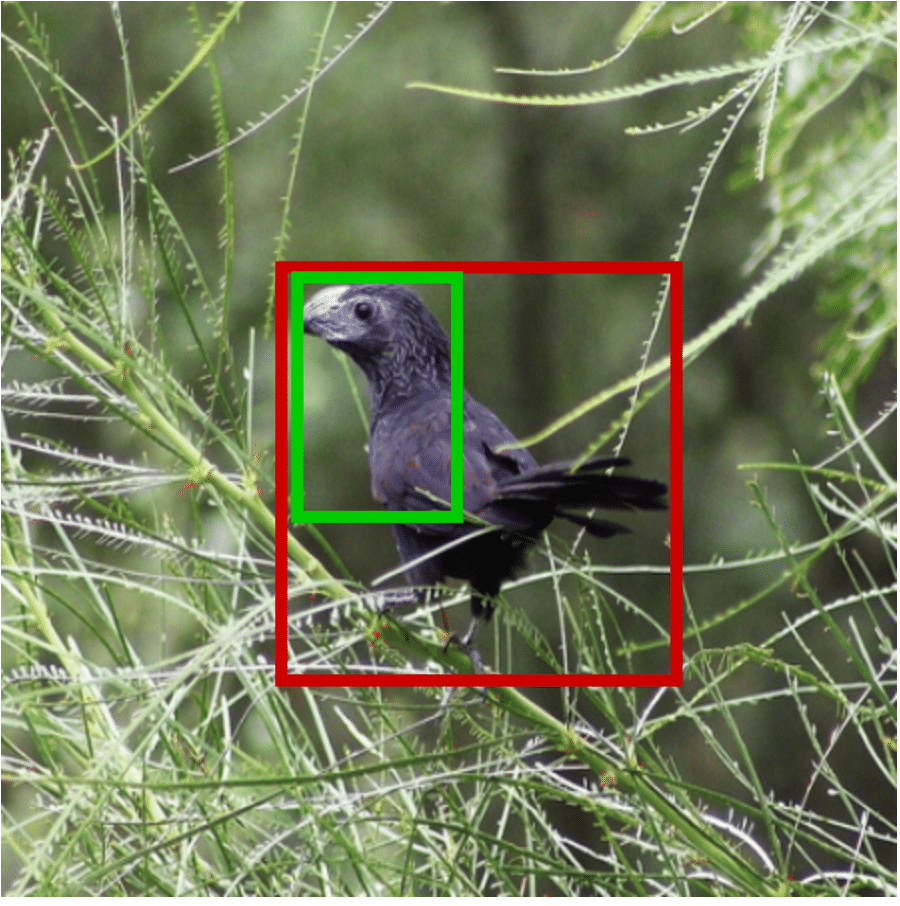}\\  (b)}&
\shortstack{\includegraphics[width=0.33\linewidth]{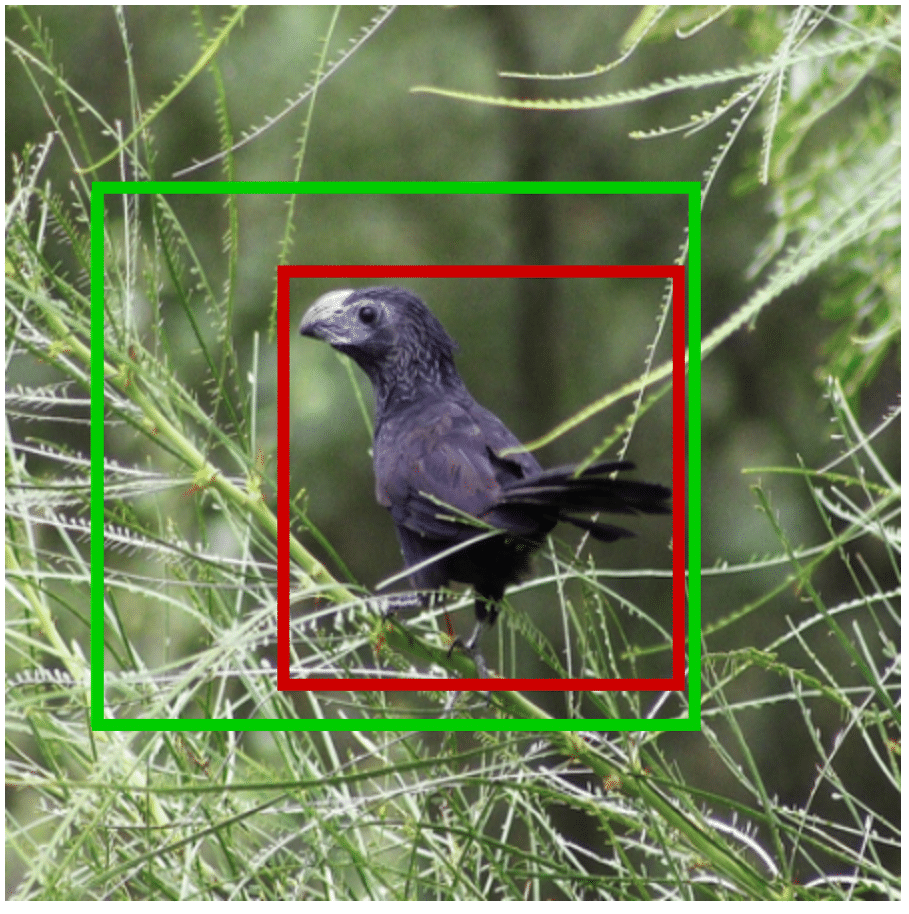}\\  (c)}&
\shortstack{\includegraphics[width=0.33\linewidth]{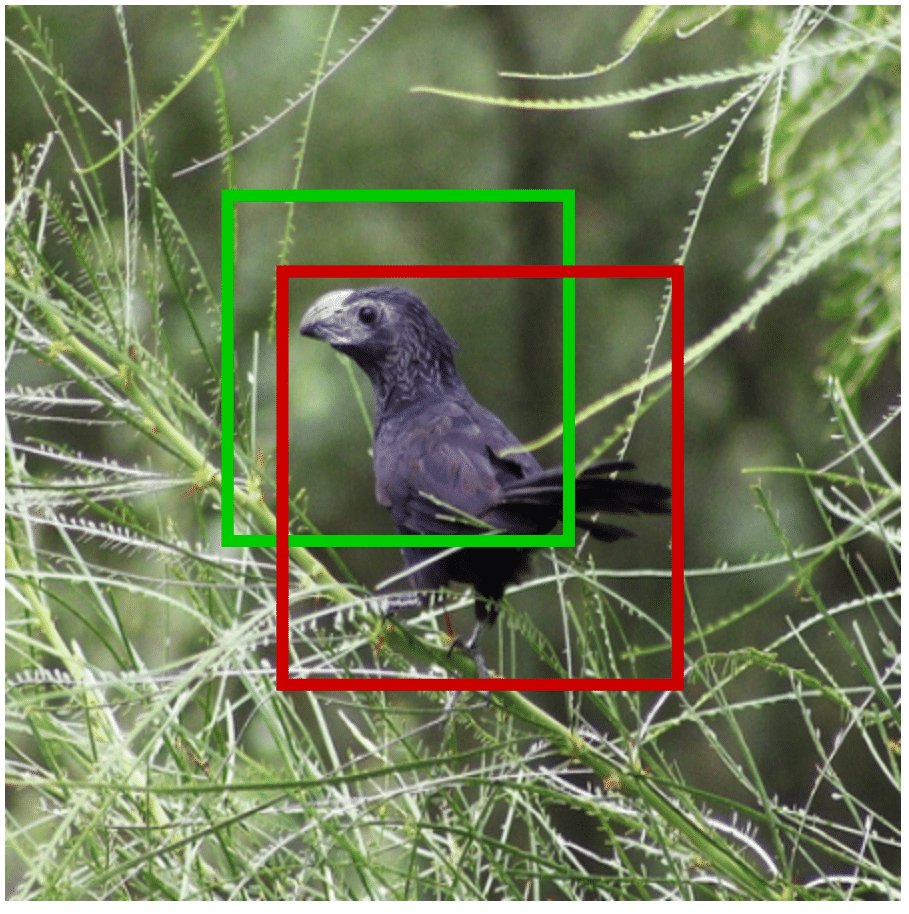}\\  (d)}\\[1pt]
\end{tabular}
}
\caption{The visualization of localization results in images between attention-guided localization, random localization, and SAC localization. Note that (a) are Inputs, (b) shows Localization results with SAC, (c) visualizes the Localization of Attention Cropping~\cite{hu2019WS_DAN}, and (d) indicates Localization results of Random Cropping~\cite{hu2018weaklyBAN}
The red-colored bounding boxes denote the ground-truth. The green-colored bounding boxes denote the predictions. Best viewed in color.}
    \label{fig:img_crop}
\end{figure}

\textbf{Localization Analysis}
Under the Joint Embedding module, the attention distribution map $\mathcal{M}$ contains informative regions used for classification, which can be leveraged to extract image-based object localization. Figure \ref{fig:img_crop} shows the comparison between Attention-guided localization introduced in  \cite{hu2019WS_DAN}, Random localization \cite{hu2018weaklyBAN},  and our SAC localization, to demonstrate the effectiveness of the localization method in Algorithm~\ref{alg:localization} of SAC. Different from Random localization, which might erase the informative regions of the object out of the image and the Attention-guided localization, which erases the background and keeps the object in the image, SAC localization can efficiently identify the informative regions which are useful for fine-grained classification between hard-to-distinguish classes. 

\begin{figure}[!ht]
  \centering
    \subfigure[]{\includegraphics[width=0.185\linewidth, height=1.3\linewidth]{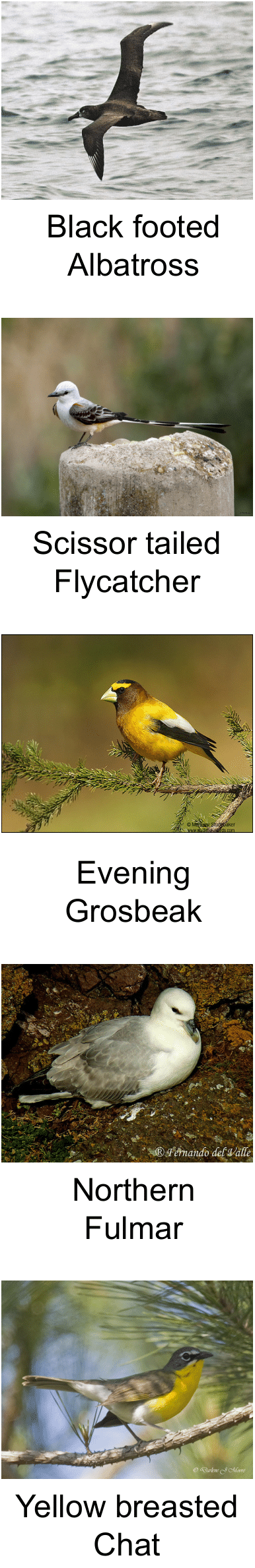}}
    \subfigure[]{\includegraphics[width=0.185\linewidth, height=1.3\linewidth]{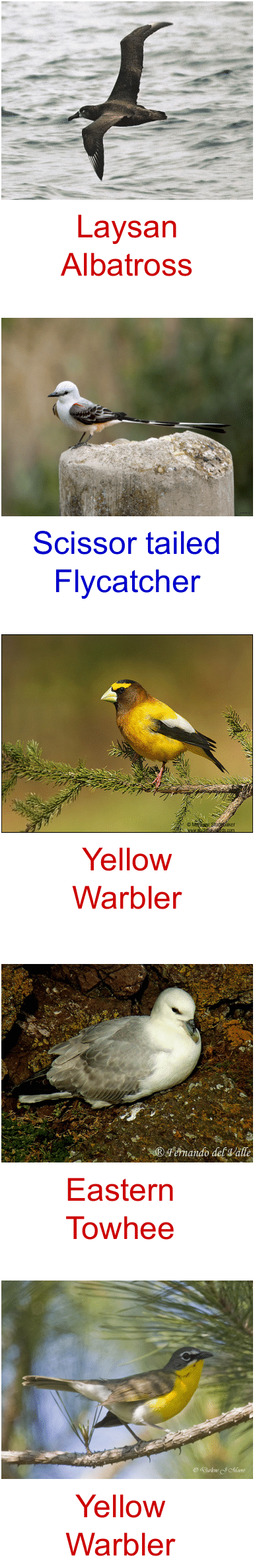}}
    \subfigure[]{\includegraphics[width=0.185\linewidth, height=1.3\linewidth]{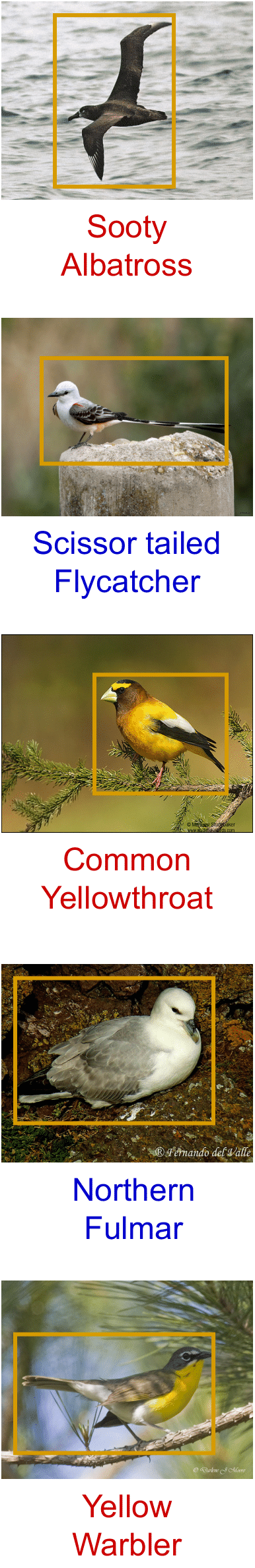}}
    \subfigure[]{\includegraphics[width=0.185\linewidth, height=1.3\linewidth]{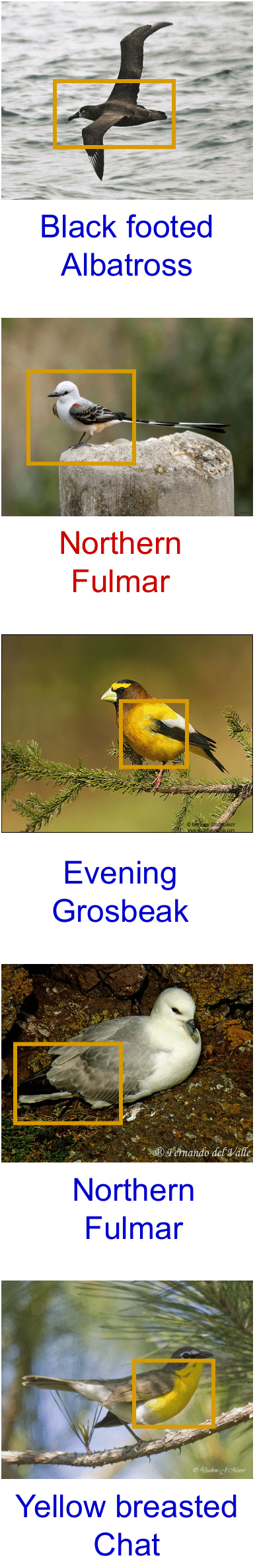}}
    \subfigure[]{\includegraphics[width=0.185\linewidth, height=1.3\linewidth]{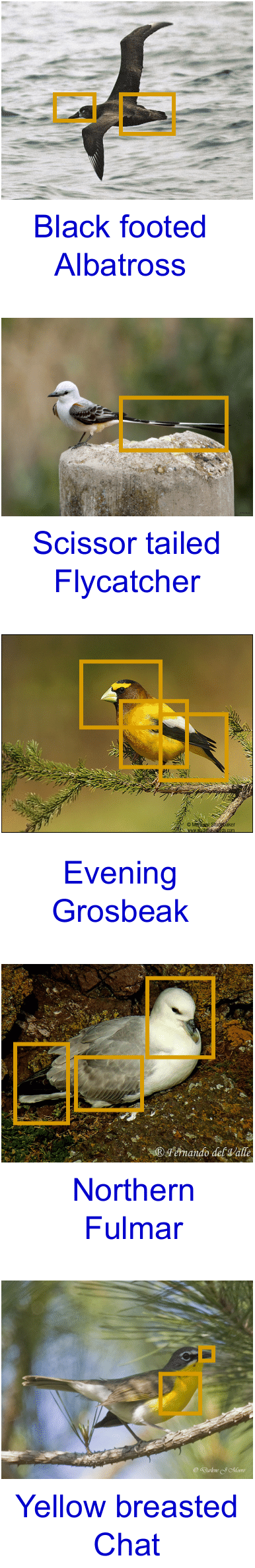}}
 \caption{Qualitative comparison of different classification methods. (a) Input image and its corresponding ground-truth label, (b) ResNet-50~\cite{he2016deepResnet152}, (c) WS\_DAN~\cite{hu2019WS_DAN}, (d) MMAL~\cite{zhang2021MMAL}, and (e) Our SAC. Boxes are localization areas. Red color indicates wrong classification result. Blue color indicate correct predicted label. Best viewed in color.}
 \label{fig:classify_vis}
\end{figure}

\textbf{Qualitative results of different classification methods.}
Fig.~\ref{fig:classify_vis} illustrates the classification results and corresponding localization areas of different methods. In all samples, we can figure out that ResNet-50 do not focus on any specific area during classify process; WS\_DAN is trained to focus on objects. MMAL and SAC are trained to focus on different specific areas. Through illustration, we can see that ResNet-50 does not show good enough results for classifying fine-grained samples. In some cases, focusing on specific object does not make WS\_DAN to give correct predictions. MMAL achieves good results in most cases, however, if the method focus on wrong areas, the model has high chance to give out wrong prediction. Self-Assessment mechanism in SAC allows it to focus on different areas based on different hard-to-distinguish classes. Thus, the method can focus on more meaningful areas and also ignore unnecessary ones. Hence, SAC achieves good predictions even with hard cases.

\section{Discussion}
\subsection{Limitation.} Our method contains different modules which are specifically designed for the fine-grained classification task. This is also a common problem of many fine-grained approaches since this task requires paying attention at very detailed levels. Based on the design, our method has three main hyper-parameters. Fortunately, the experiments show that we can achieve good results with a unified setup. Therefore, there is no need to tune these parameters further. Compared to other approaches, our method utilizes the class name as the input. This is to provide more supervised information during training. While this arguably brings more information, we note that the class name is identical with the class identity, which is used by all methods to calculate the loss. In practice, as we add an extra self assessment classifier to the backbone, our method increases the training time and testing time by approximately $10\%$. We believe that this is a reasonable trade-off to improve the classification accuracy by $1.2\%$. Finally, while in theory we can integrate our method to any other fine-grained classifiers, we note that it is not straightforward to integrate our proposed method to multi-step approaches such as Parts Models~\cite{ge2019weakly3stepmodelling}. This is because these methods do not allow end-to-end training, hence making the integration more tedious. 

\subsection{Broader Impact.}
We have proposed a new fine-grained classification method that significantly improves the current state of the art. More efficient fine-grained methods will have direct impacts on different domains in real-world problems such as crop disease detection. We hope that our method will have broader impacts by enabling the integration into any new fine-grained classifiers in the future. 

\section{Conclusion}
We introduce a Self Assessment Classifier (SAC) which effectively learns the discriminative features in the image and resolves the ambiguity from the top-k prediction classes. Our method generates the attention map and uses this map to dynamically erase unnecessary regions during the training. The intensive experiments on CUB-200-2011, Stanford Dogs, and FGVC Aircraft datasets show that our proposed method can be easily integrated into different fine-grained classifiers and clearly improve their accuracy. In the future, we would like to test the effectiveness of our method on more classifiers such as EfficientNet \cite{tan2019efficientnet} and bigger datasets such as ImageNet \cite{deng2009imagenet}.

{\small
\bibliographystyle{ieee_fullname}
\bibliography{egbib}
}
\end{document}